\documentclass[journal]{IEEEtran}
\usepackage{booktabs}
\usepackage{multirow}
\usepackage{amsmath,amsfonts}
\usepackage{array}
\usepackage[caption=false,font=normalsize,labelfont=sf,textfont=sf]{subfig}
\usepackage{textcomp}
\usepackage{stfloats}
\usepackage{accents} 
\usepackage{url}
\usepackage{verbatim}
\usepackage{amsmath}
\usepackage{graphicx}
\usepackage{balance}
\usepackage{pgfplots}
\pgfplotsset{compat=1.18}
\usepackage{algorithm}  
\usepackage{algpseudocode} 
\usepackage{amssymb}
\usepackage{pifont}
\usepackage[colorlinks=true, linkcolor=blue, citecolor=blue, urlcolor=red]{hyperref}
\algnewcommand{\LeftComment}[1]{\Statex \(\triangleright\) #1}

\usepackage{soul}    %
\usepackage{xcolor}  %
\usepackage[table]{xcolor} %
\soulregister\cite7
\soulregister\ref7
\soulregister\eqref7
\soulregister\mathbf7

\begin{document}

\title{Prompt to Restore, Restore to Prompt: Cyclic Prompting for Universal Adverse Weather Removal}
\author{Rongxin Liao,
    Feng~Li,
    Yanyan Wei,
    Zenglin Shi,
    Le Zhang,
    Huihui Bai,~\IEEEmembership{Senior Member,~IEEE}
	and~Meng~Wang~\IEEEmembership{Fellow,~IEEE}

\thanks{
This work was supported in part by Beijing Natural Science Foundation (L223022), in part by National Natural Science Foundation of China (62331003, 62302141, 62472138, U24A20331), and in part by Anhui Provincial Natural Science Foundation (2408085MF159).
}
\thanks{
Rongxin Liao, Feng Li, Yanyan Wei, Zenglin Shi, and  Meng Wang are with the School of Computer Science and Information Engineering, Hefei University of Technology, Hefei, Anhui 230601, China (e-mail: rongxinliao@mail.hfut.edu.cn; fengli@hfut.edu.cn; weiyy@hfut.edu.cn; zenglin.shi@hfut.edu.cn;wangmeng@hfut.edu.cn).

Le Zhang is with the School of Information and Communication Engineering, University of Electronic Science and Technology of China, Chengdu, Sichuan 611731, China (email: zhangleuestc@gmail.com).

Huihui Bai is with the School of Computer Science and Technology, Beijing Jiaotong University, Beijing 100044, China (email: hhbai@bjtu.edu.cn).

Corresponding author: Feng Li
}}
\maketitle


\begin{abstract}
Universal adverse weather removal (UAWR) seeks to address various weather degradations within a unified framework. Recent methods are inspired by prompt learning using pre-trained vision-language models (\emph{e.g.}, CLIP), leveraging degradation-aware prompts to facilitate weather-free image restoration, yielding significant improvements. In this work, we propose CyclicPrompt, an innovative cyclic prompt approach designed to enhance the effectiveness, adaptability, and generalizability of UAWR. CyclicPrompt comprises two key components: 1) a composite context prompt that integrates weather-related information and context-aware representations into the network to guide restoration. This prompt differs from previous methods by marrying learnable input-conditional vectors with weather-specific knowledge, thereby improving adaptability across various degradations. 2) The erase-and-paste mechanism, after the initial guided restoration, substitutes weather-specific knowledge with constrained restoration priors, inducing high-quality weather-free concepts into the composite prompt to further fine-tune the restoration process. Therefore, we can form a cyclic ``Prompt-Restore-Prompt'' pipeline that adeptly harnesses weather-specific knowledge, textual contexts, and reliable textures. Extensive experiments on synthetic and real-world datasets validate the superior performance of CyclicPrompt. The code is available at: \href{https://github.com/RongxinL/CyclicPrompt}{https://github.com/RongxinL/CyclicPrompt.}  

\end{abstract}


\section{Introduction}
\IEEEPARstart
{I}{mages} captured in adverse weather, such as rain, snow, and fog/haze, often suffer from diminished quality, which presents significant challenges for outdoor vision systems~\cite{Ithaca365,reddy2022master,weatherstream}. Most methods primarily focus on crafting specialized networks tailored to certain weather conditions, including image deraining~\cite{li2023dilated,chen2023learning, wei2021deraincyclegan}, defogging/dehazing~\cite{dehaze-sam,guo2022image}, and desnowing~\cite{Dataset-Snow100K}. Some other studies~\cite{Restormer,Uformer,GRL} have developed a general restoration architecture and trained discrete copies independently to accommodate different weather conditions. Although notable outcomes have been obtained in their intended scenarios, these task-specific models struggle to generalize effectively across different weather types and even intensity levels. This implies that tedious specialized weights are necessitated in practical applications, increasing computational costs and resources. Chen~\emph{et al.}~\cite{chen2021ipt} develop pre-trained image processing models comprising multiple heads aligned to different low-level tasks, which excavates the model capability to deal with different corruptions but requires further fine-tuning on each specific dataset. 

\begin{figure*}
	\centering
    \includegraphics[width=1\linewidth]{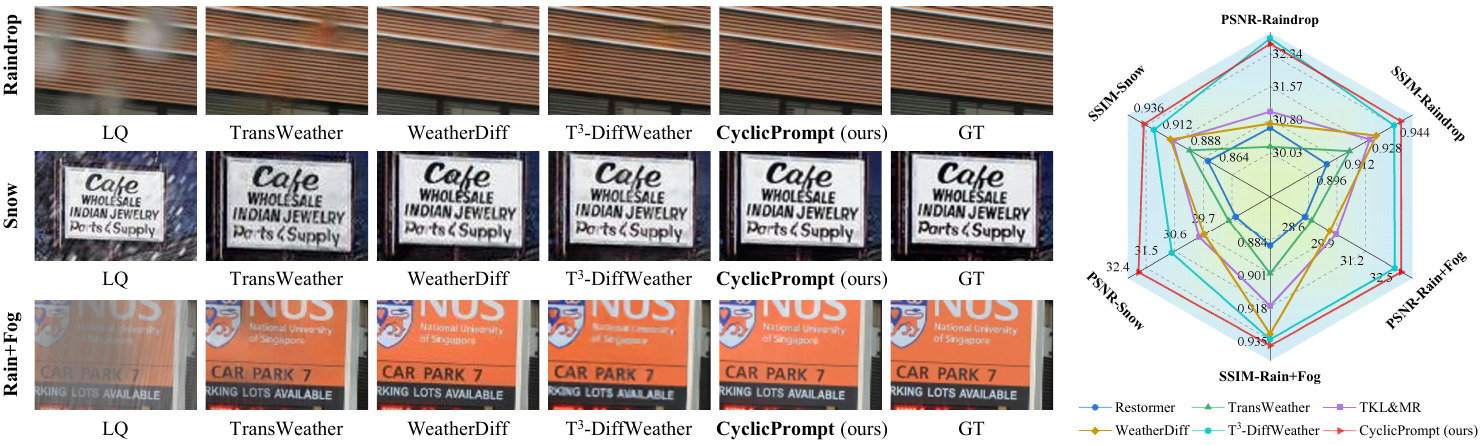}
	\caption{Our CyclicPrompt is capable of removing multiple adverse weather degradations in one model and producing more visually pleasant images than several state-of-the-art UAWR methods: TransWeather\cite{TransWeather}, WeatherDiff\cite{WeatherDiff}, and T$^3$-Diffweather\cite{T3-DiffWeather}. The quantitative comparisons in PSNR and SSIM also indicate the preferable performance of our method obviously.}
	\label{fig:teaser}
\end{figure*}

Recent studies~\cite{Dataset-AllWeather, TransWeather, ADFSD, TUM_chen, domainMPR, AWRCP} have shifted towards developing generic solutions to address multiple adverse weather challenges within a single, unified model. 
Specifically, Li~\emph{et al.}~\cite{Dataset-AllWeather} derive multiple encoders to learn the weather-specific representations and use the neural architecture search (NAS) technique to find the optimal one for each degradation. TransWeather~\cite{TransWeather} utilizes a shared encoder-decoder with weather-type queries to implicitly modulate degradation patterns. \cite{TUM_chen,domainMPR} follows the multi-encoder design in \cite{Dataset-AllWeather} to accommodate diverse degradations. Although some generalizability improved on synthetic datasets, their adaptability to real-world scenarios remains limited. Inspired by the power of large-scale pre-trained vision-language models (PVLM)~\cite{CLIP} in various computer vision tasks~\cite{LISA, SD}, some works leverage off-the-shelf PVLMs to extract degradation-aware prompts in linguistic~\cite{DA-CLIP, CVPR24-Languagedriven} or visual~\cite{tan_tip24,T3-DiffWeather} domains and interact with the main feature flow hierarchically. For instance, DA-CLIP~\cite{DA-CLIP} aligns text embeddings with visual representations to predict the degradation prompt integrated into other restoration models. Liang~\textit{et al.}~\cite{CVPR24-Languagedriven} construct a question prompt based on pre-trained LISA~\cite{LISA} to estimate the weather types and levels. Despite remarkable advancements, existing prompt-driven methods for UAWR still fall short in several limitations: 1) While PVLMs play a pivotal role in navigating weather-related content, describing weather types through simple text or converted image embeddings remains challenging, particularly in scenarios with complex weather (\textit{e.g.}, rain+fog, Fig.~\ref{fig:teaser}). 2) Visual prompts from corrupted images may not adequately express the spatial context or even introduce ambiguous semantics due to the interference of weather degradations.

To tackle these issues, this paper presents an in-depth investigation into the construction of effective prompts to fully unleash the potential of prompt learning. To this end, we propose a remarkably different cyclic prompt learning approach for UAWR, termed CyclicPrompt, which integrates weather-specific knowledge, textual context, and reliable textures, so that better results can be yielded. Specifically, we construct a composite context prompt (C2P) to perceive weather-related information and context-aware representations. The first component of C2P aligns with existing methods that extract CLIP-based visual features but is further enhanced by incorporating learnable input-conditional vectors, enabling dynamic adaptation of weather-specific knowledge to improve adaptability. The second is derived from explicit textual descriptions to mitigate the semantic ambiguities caused by adverse weather. While the composite prompt shows evident effects, it may
not sufficiently facilitate more realistic texture recovery as it primarily focuses on compact semantics rather than visual characteristics. To overcome this, after the C2P-driven UAWR, we propose the erase-and-paste mechanism (EPM). It leverages constrained restoration priors to substitute the weather-related part with high-quality weather-free concepts in C2P, strengthening the information flow in a cyclic “Prompt-Restore-Prompt” process. On the other hand, along with EPM, we also devise the residual prior modulator based on restored results to provide plausible texture clues and structural appearance with different granularities. 

In summary, the main contributions of our work are:
\begin{itemize}
    \item We present CyclicPrompt, a distinct cyclic prompt-driven universal framework that effectively and adaptively removes weather degradations in an all-in-one manner for UAWR. Extensive experiments on both synthetic and real-world datasets demonstrate its superiority.
    \item We construct a composite context prompt (C2P) that combines CLIP-equipped weather-specific knowledge, learnable input-conditional vectors, and textual prompt to capture weather-related information and context-aware representations. 
    \item We introduce the erase-and-paste mechanism (EPM), leveraging constrained restoration priors to embed weather-free representations into the C2P, thereby refining the restoration process. Additionally, the residual prior modulator is designed to enrich visual characteristics for better detail recovery along with EPM.
\end{itemize}

The rest of the paper is organized as follows. Section~\ref{sec:related work} reviews the related works. Section~\ref{sec:method} introduces the proposed CyclicPrompt in detail. Section~\ref{sec:experiments} shows the experimental results of our method and validates the effects of the proposed components. In Section~\ref{sec:conclusion}, we conclude this work.

\section{Related Work}
\label{sec:related work}
\subsection{Task-Specific Adverse Weather Removal}
Task-specific methods~\cite{li2023dilated,guo2022image,Restormer,Uformer} train the individual model for each specific weather type like rain~\cite{li2023dilated,chen2023learning,Dataset-RainDS,Dataset-Raindrop}, snow~\cite{chen2021all,Dataset-Snow100K,DDMSNet}, and fog~\cite{guo2022image,wu2023ridcp,HRGAN,Pix2Pix}. In literature, tremendous methods are tailored to these degradations, known as image draining, dehazing, and desnowing \emph{et al.}. Here, we provide a brief review of them.

\textbf{Image Deraining}. Rainy weather usually affects the images in two ways: rain streaks and raindrops. Fu~\emph{et al.}~\cite{fu2017clearing} propose the first convolutional neural network (CNN)-based method DerainNet for rain streak removal. As motivated, a series of methods are developed that utilize advanced deep networks~\cite{Dataset-Rain100H,li2018recurrent,RainGAN-ID-CGAN,deng2020detail,HazeCNN-GCANet, RainCNN-PReNet,Rain-Rcp} or rain-related physical modeling~\cite{HRGAN,wang2020model,pan2020physics,yu2023both} to pursue promising results. For instance, \cite{Dataset-Rain100H,li2018recurrent} adopt recurrent networks to progressively remove the rain by modelling the feature dependencies across multiple recurrent phases. \cite{Rain-Rcp} and \cite{li2018robust} explore the residue channel prior to learn rain-free object structures for rain removal. Pan~\emph{et al.} propose a physics-based GAN model to ensure that the restored image is consistent with the corrupted observation. \cite{yu2023both} investigates the physical relationship of the rain attribute and style between synthetic and real data.

To alleviate the limitation of CNNs in long-range dependency capturing, \cite{chen2023learning,RainTransformer-IDT,wang2023multi} combines the locality of CNNs and the global capacity of transformers, which performs favorably. Raindrops adhere to glass surfaces (\emph{e.g.} windows or camera lens) and occlude the regions of background scenes. To mitigate the damage of visibility, a popular approach~\cite{quan2019deep,Dataset-Raindrop,shao2021uncertainty} is to detect the raindrop masks from rainy inputs to help recover raindrop-free ones. There are also several methods~\cite{Dataset-RainDS,zhang2021dual} jointly remove the rain streaks and raindrops, significantly improving the applicability in real rainy scenes. 

\textbf{Image Dehazing}. Hazy images often suffer from color distortion, blurry textures, and poor quality. In past years, the prevailing progress of image dehazing has been witnessed in deep learning~\cite{cai2016dehazenet,li2017aod,zhang2018densely,HazeGAN-PMHLD,qin2020ffa,dong2020multi}. DehazeNet~\cite{cai2016dehazenet} introduces the pioneering CNN for haze removal. DCPDN~\cite{zhang2018densely} presents an edge-preserving densely connected pyramid network to estimate the transmission map in image dehazing. FFA-Net~\cite{qin2020ffa} devises the feature attention to facilitate the flexibility in dealing with different information. Considering that the haze on objects is highly related to depth, recent research~\cite{yang2022self,wang2024selfpromer,zhang2024depth} recover from the hazy image by depth estimation and depth-aware reconstruction. Tian~\emph{et al.}~\cite{ye2024learning} integrate diffused texture priors for high-quality image restoration, which has demonstrated powerful generative ability. 

\textbf{Image Desnowing}. Unlike rain or haze, the particle shapes and sizes in snow introduce greater complexity. From the first deep learning method DesnowNet~\cite{Dataset-Snow100K}, deep learning has dominated this task~\cite{chen2021all,JSTASR,liang2022drt,chen2023uncertainty}. DDMSNet~\cite{DDMSNet} learns semantic- and geometry-aware representations utilizing semantic and depth priors. JSTASR~\cite{JSTASR} performs joint size and transparency-aware desnowing using partial convolutions and veiling effect removal. In~\cite{chen2023cplformer}, Chen~\emph{et al.} propose the CPLFormer that learns from cross-scale snow prototypes to handle complicated snow landscapes. Zhang~\emph{et al.}~\cite{zhang2023hcsd} propose the hybrid color space-based desnowing network based on the analysis of snowflake noises in the Hue channel. While the above studies acquire impressive achievements on specific weather conditions, they ignore other adverse weather which limits their scope of practical applications.

\subsection{Universal Adverse Weather Removal}
UAWR~\cite{TransWeather,AirNet} aims to develop a single model to recover weather-free images under different adverse weather, which is a long-standing problem in the computer vision community. Many of them aggregate parallel weather-specific encoders~\cite{Dataset-AllWeather,domainMPR,TUM_chen} or decoders~\cite{WGWS} where each learns certain weather representations in a unified architecture for multi-weather removal. AWRCP~\cite{AWRCP} utilizes the codebook priors by pre-trained VQGAN~\cite{VQGAN} from undistorted images to improve recovered details. Motivated by the generative capacity of diffusion models~\cite{DDPM,SD}, WeatherDiff~\cite{WeatherDiff} presents a patch-based diffusive algorithm, achieving promising generalization ability on synthetic and real-scenario multi-weather restoration. In more recent years, prompt learning with pre-trained models~\cite{CLIP,LISA} has emerged as a powerful tool to learn degradation-aware information effectively from degraded images~\cite{DA-CLIP,tan_tip24,CVPR24-Languagedriven}, thereby empowering the effectiveness and generalizability of restoration models. This work introduces a cyclic prompt learning framework for UAWR, which shows the superiority of various weather degradations.

\subsection{Prompt Learning} 
Prompt learning originates from natural language processing (NLP)~\cite{bert,GPT}, which provides in-context information to adapt models to other tasks with minimal fine-tuning~\cite{lester2021power,liu2021p}, particularly benefiting large-scale pre-trained models. Inspired by the pioneering attempt CLIP~\cite{CLIP} that applies PVLMs to downstream tasks, prompt learning has been expanded in computer vision domains~\cite{jia2021scaling,jia2022visual,bahng2022visual}, which incorporates handcrafted prompt templates or learnable prompt tokens to guide the interaction between textual and visual tokens. CoOp~\cite{Coop} models a prompt's context words with learnable vectors to adapt CLIP-like vision-language models for downstream image recognition. CoCoOp~\cite{cocoop} further introduces conditional context optimization that learns input-conditional dynamic prompts with the static prompts in CoOp~\cite{Coop} to improve its transferability. Building upon these advancements, researchers have also explored the advantages of prompt learning for low-level vision tasks. CLIP-LIT~\cite{CLIP-Lit} devises iterative prompt learning to characterize backlit and well-lit images more precisely. SeeSR~\cite{Seesr_CVPR24} generates image tag text as a semantic prompt for real-world image super-resolution. This work exploits prompt learning but induces significant insights to achieve effective and generalized UAWR.

\begin{figure*}
    \centering
    \includegraphics[width=0.92\linewidth]{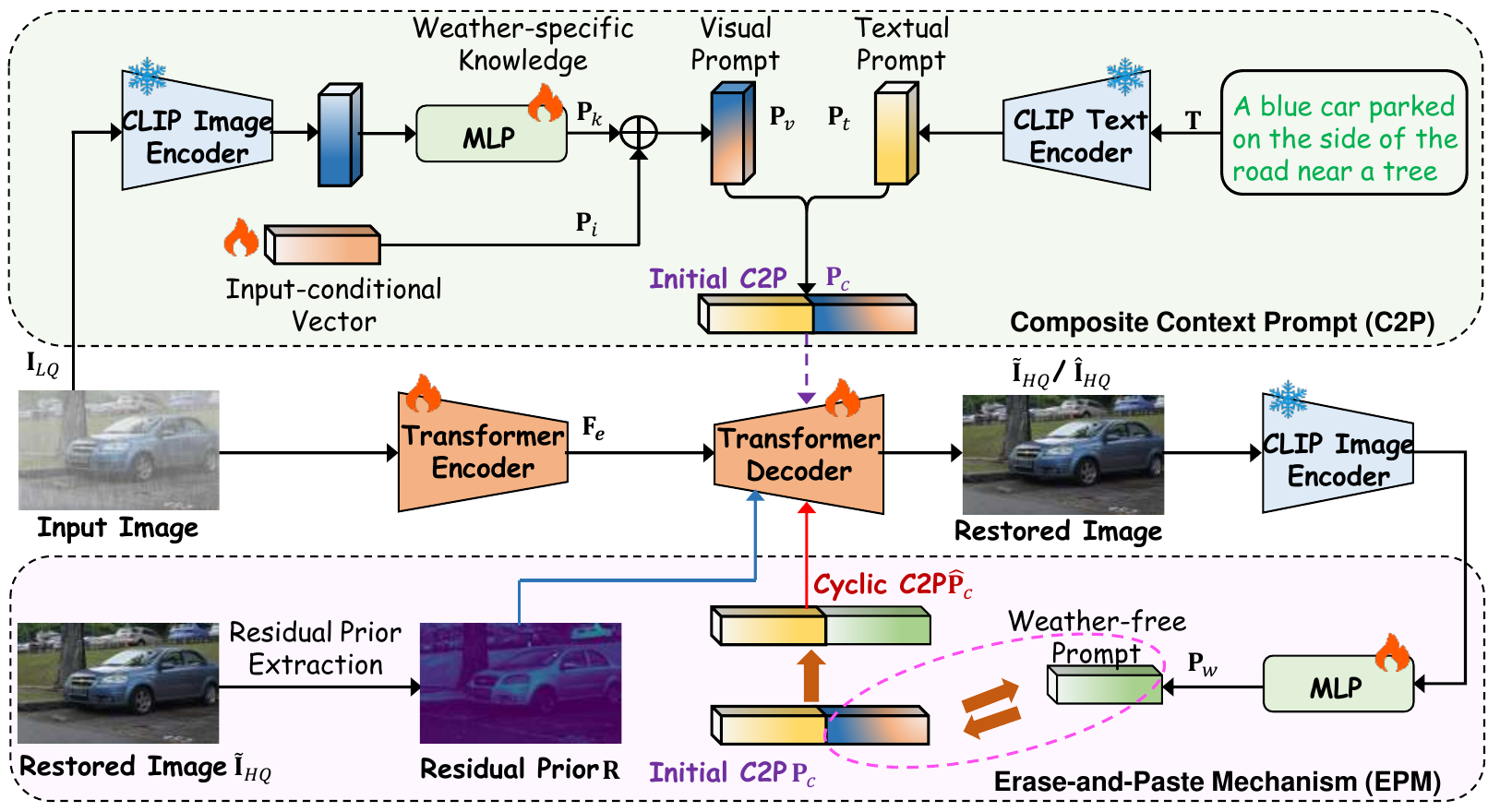}
    \caption{Overview of the proposed CyclicPrompt built upon a transformer architecture following a cyclic “Prompt-Restore-Prompt” process for UAWR. We first construct a composite context prompt (C2P) containing weather-specific knowledge, input-conditional vector, and textual prompt to perceive weather-related information and context-aware representations. After the C2P-driven UAWR, the erase-and-paste mechanism (EPM) leverages constrained restoration priors to substitute the weather-related part in C2P with the weather-free prompt to facilitate the decoder flow further. The residual prior is extracted to provide texture clues and structural appearance for better detail recovery.} 

    \label{fig:main}
\end{figure*}

\section{Method}
\label{sec:method}

\subsection{Overview}
\label{sec:overview}
 The overall framework of the proposed CyclicPrompt is illustrated in Fig. \ref{fig:main}. The core components are 1) the composite context prompt (C2P) capturing weather-related information and context-aware textual representations (Sec.~\ref{sec:c2p}), and 2) the erase-and-paste mechanism (EPM) providing weather-free representations into the C2P for further refinement (Sec.~\ref{sec:epm}). We build the prompt block to adapt the deep image features based on our C2P (Sec.~\ref{sec:rpm}). Moreover, we also extract the residual prior to improve the detail recovery (Sec.~\ref{sec:rpm}). Our CyclicPrompt establishes a ``Prompt-Restore-Prompt'' pipeline with the C2P and EPM.

Given a low-quality image \(\mathbf{I}_{LQ}\) degraded by unknown adverse weather, the goal of our CyclicPrompt is to learn a universal model $\mathcal{M}$ and restore a photo-realistic weather-free image. We build the overall framework upon a U-shaped transformer architecture. Generally, the transformer encoder first maps the high-resolution input $\mathbf{I}_{LQ}$ into a lower dimensional latent space via $\mathbf{F}_{e}=\mathcal{E}(\mathbf{I}_{LQ})$. The transformer decoder gradually recover the clean output $\mathbf{I}_{HQ}$ from $\mathbf{F}_{e}$. We incorporate C2P $\mathbf{P}_c=\{\mathbf{P}_v,\mathbf{P}_t\}$ in the decoder part via the prompt block (Fig.~\ref{fig:transformer}) to guide the decoding process. After the initial restoration, the EPM replaces the weather-related information $\mathbf{P}_v$ with the weather-free prompt $\mathbf{P}_w$ extracted from $\tilde{\mathbf{I}}_{HQ}$, constituting an enhanced prompt $\hat{\mathbf{P}}_c$ to strengthen the information flow. Simultaneously, we also extract the residual prior $\mathbf{R}$ from $\tilde{\mathbf{I}}_{HQ}$, which is then equipped with a residual prior modulator (RPM) in the prompt block to provide detail representations for restoration fidelity. We take $\mathbf{R}$ and $\hat{\mathbf{P}}_c$ as the guidance to facilitate the decoding process further, endowing the model with more promising generative capacity and thus producing the final reconstruction $\hat{\mathbf{I}}_{HQ}$. 

In CyclicPrompt, both two restoration processes are constrained by L1 loss. Therefore, the total loss function for training our model can be formulated as


\begin{equation}
    \mathcal{L} = \Vert \tilde{\mathbf{I}}_{HQ}-\mathbf{I}_{HQ} \Vert_1 + 
    \lambda \Vert \hat{\mathbf{I}}_{HQ} - \mathbf{I}_{HQ} \Vert_1 , 
\end{equation}
where $I_{HQ}$ denotes the ground-truth image of $\mathbf{I}_{LQ}$ 
and $\lambda$ is the hyper-parameter that controls the impact of the second iteration. We empirically set $\lambda=1$ in our experiments.

\subsection{Composite Context Prompt}
\label{sec:c2p}
In this work, we consider the prompt in the UAWR task from two aspects to build the C2P $\mathbf{P}_c$. First, we integrate weather-specific knowledge $\mathbf{P}_k$ and learnable input-conditional vector $\mathbf{P}_i$ as the visual prompt $\mathbf{P}_v$ to perceive the weather-dedicated information. Second, the text description for the input image can be easily accessible to encapsulate the semantic information, serving as the textual prompt $\mathbf{P}_t$.

Recent works~\cite{DA-CLIP,Seesr_CVPR24} have demonstrated the potential of text representations in image restoration as it is easily accessible and intuitively expresses the semantics of the objects, environment, and contexts in an image. In this work, for the input image $\mathbf{I}_{LQ}$, we apply BLIP~\cite{BLIP} to generate corresponding caption $\mathbf{T}$. Then, considering CLIP~\cite{CLIP} can effectively encode the correlations between visual and linguistic data, we obtain the textual prompt $\mathbf{P}_t$ by a frozen CLIP text encoder as follows:
\begin{equation}
   \mathbf{P}_t = \mathcal{E}_{clip}^{text}(\mathbf{T}),
\end{equation}
where $\mathcal{E}_{clip}^{text}$ denote the text encoding operation.  $\mathbf{P}_t\in\mathbb{R}^{1\times D}$ and $D$ is the channel dimension.

In real scenarios, where complex degradation issues are often encountered, accurately modeling the degradation through text descriptions remains a significant challenge. To alleviate this limitation, we exploit the frozen CLIP image encoder to transform $\mathbf{I}_{LQ}$ into deep features and perceive weather degradations, which are then processed by a multilayer perceptron (MLP), yielding the weather-specific knowledge $\mathbf{P}_k\in\mathbb{R}^{1\times D}$
\begin{equation}
    \mathbf{P}_k ={\rm{MLP}}_1( \mathcal{E}_{clip}^{img}(I_{LQ})),
\end{equation}
where $\mathcal{E}_{clip}^{img}$ denotes the image transformation function. To improve the adaptability of models, inspired by the conditional prompt learning in CoOp~\cite{Coop}, we incorporate a learnable input-conditional vector $\mathbf{P}_i$ for the CLIP-equipped image feature to dynamic calibrate and adapt the weather-dedicated information to complex unknown weather conditions, which is combined with $\mathbf{P}_k$ to generate the visual prompt
\begin{equation}
    \mathbf{P}_v = \mathbf{P}_k + \mathbf{P}_i,
\end{equation}
where \(\mathbf{P}_i \in \mathbb{R}^{N\times D}\) is random initializing tunable vector which length is $N$. We replicate $\mathbf{P}_k$ to $N$ copies, thus being the same size with $\mathbf{P}_v$.

Based the above discussion, the initial C2P $\mathbf{P}_c\in\mathbb{R}^{(1+N)\times D}$ can be constructed by concatenating the weather-related visual prompt \(\mathbf{P}_v\) and the textual prompt \(\mathbf{P}_t\) 
\begin{equation}
    \mathtt{Initial:}\quad \mathbf{P}_c = {\rm{Concat}}(\mathbf{P}_v, \mathbf{P}_t).
\end{equation}

We inject C2P into the CyclicPrompt framework at the transformer decoder through prompt blocks (Fig.~\ref{fig:transformer}) to interact with the mainstream image features to improve the representations for the first guided restoration process (1st iteration) in our cyclic prompt learning framework, \emph{i.e.} ``Prompt-Restore''.

\subsection{Erase-and-Paste Mechanism}
\label{sec:epm}
With the guidance of C2P, we have the initial result $\tilde{\mathbf{I}}_{HQ}$. In contrast to existing methods~\cite{PromptIR,CVPR24-Languagedriven}, this work makes a forward step, which regards the recovered image $\tilde{\mathbf{I}}_{HQ}$ as a restoration prior to distill restoration-related information for further refinement. On the one hand, $\tilde{\mathbf{I}}_{HQ}$ is fed into another frozen CLIP image encoder followed by a trainable MLP to extract high-quality image features that are referred to the weather-free prompt $\mathbf{P}_w\in\mathbb{R}^{1\times D}$
\begin{equation}
    \mathbf{P}_w ={\rm{MLP}}_2(\mathcal{E}_{clip}^{img}(\tilde{\mathbf{I}}_{HQ})).
\end{equation}

We propose the EPM, which first erases the weather-related representations $\mathbf{P}_v$ in $\mathbf{P}_c$ and then pastes $\mathbf{P}_w$ on it, 

\begin{equation}
\begin{aligned}
  &\mathbf{P}_v \rightleftarrows \mathbf{P}_w,\\
    \mathtt{Cyclic:}\quad&\hat{\mathbf{P}}_c = {\rm{Concat}}(\mathbf{P}_w, \mathbf{P}_t),
\end{aligned}
\end{equation}
where $\hat{\mathbf{P}}_c\in\mathbb{R}^{2\times D}$ denotes the new reconstituted cyclic C2P. Therefore, there is a 2nd iteration, ``Restore-Prompt'', which applies $\hat{\mathbf{P}}_c$ as the guidance that induces weather-free concepts to optimize the restoration features, empowering the model with more effectiveness and generalizability.

\begin{figure}[t]
    \centering
    \includegraphics[width=\linewidth]{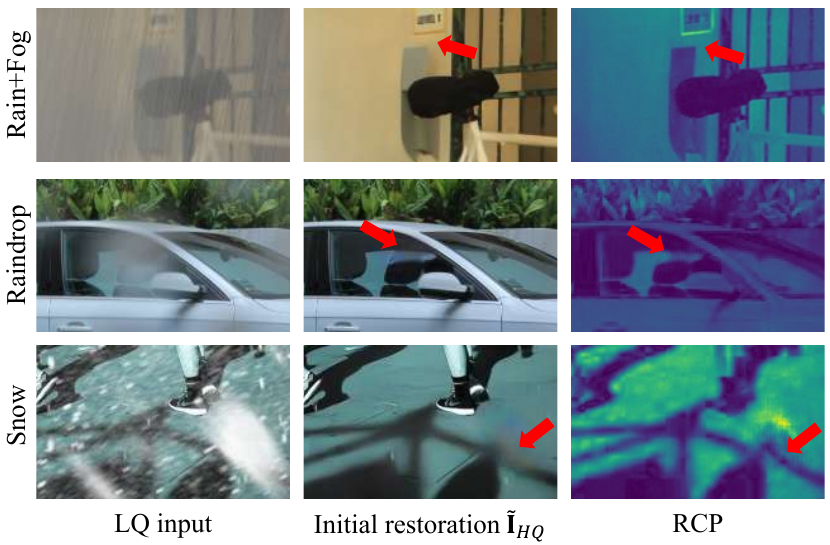}
    \caption{Illustration of the residual prior extraction. Each row corresponds to a representative sample under different adverse weather conditions. From left to right: (1) observed LQ images; (2) the initial restoration $\tilde{\textbf{I}}_{HQ}$ by C2P; and (3) the RCP extracted from $\tilde{\textbf{I}}_{HQ}$, which preserves degradation-free structural features with enhanced edge information.}
    \label{fig:vis_lq_iter0_rcp}
\end{figure}

\begin{figure*}[t]
    \centering
    \includegraphics[width=\linewidth]{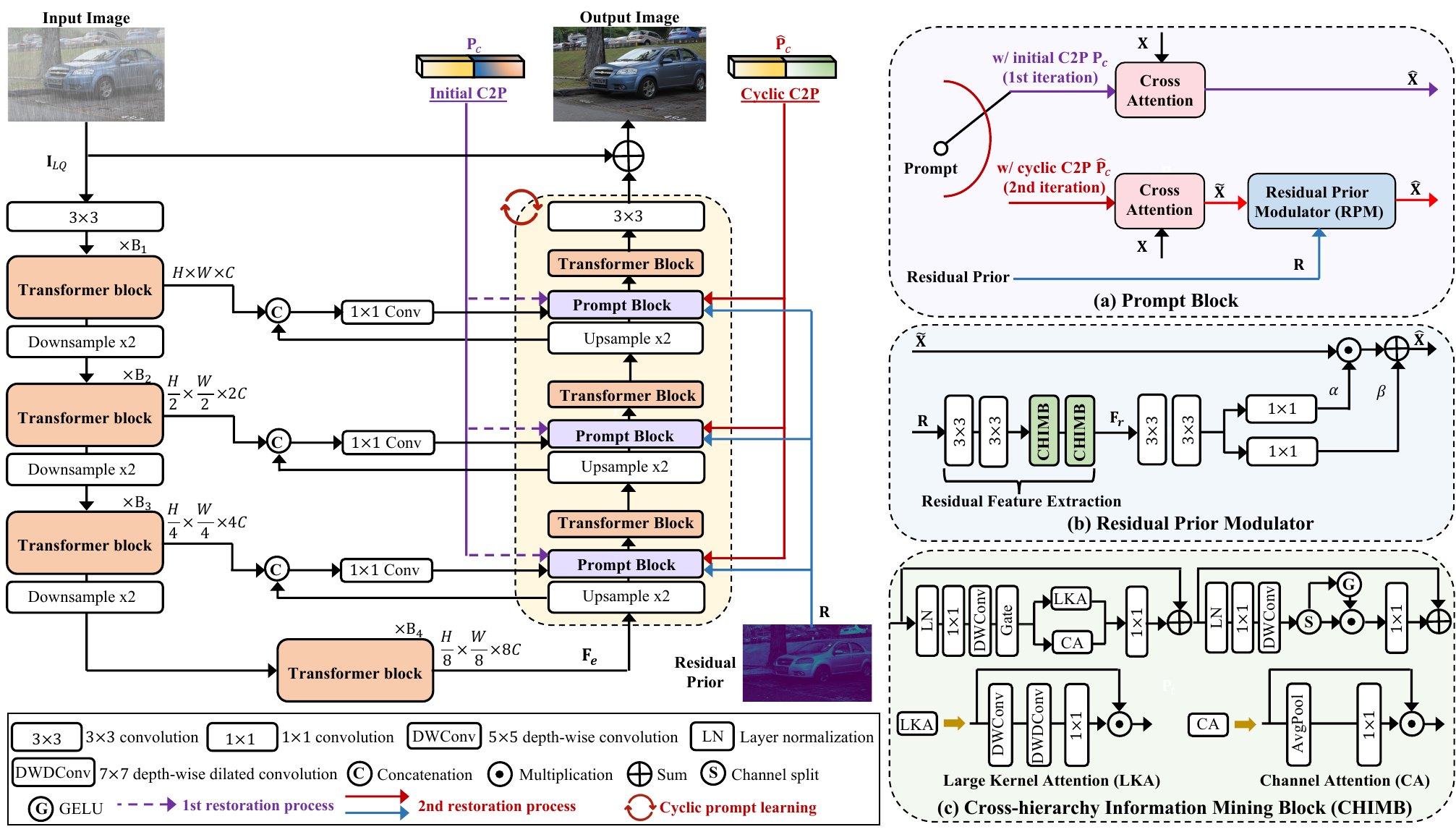}
    \caption{The 4-level transformer architecture in CyclicPrompt and each level contains several transformer blocks. The prompt block is inserted before each level of the decoder, where each contains two branches with the branch selection controlled by the restoration process: the 1st iteration with initial C2P (top branch) and the 2nd with cyclic C2P (bottom branch). We use the cross-hierarchy information mining blocks (CHIMB)~\cite{CVHSSR} to extract residual features.}
    \label{fig:transformer}
\end{figure*}

\textbf{Residual Channel Prior}. While the initial restoration in CyclicPrompt significantly reduces degradation, several severely degraded inputs still retain residual weather artifacts in their results. To address this issue, we draw inspiration from the residual channel prior (RCP) in rain removal~\cite{Dataset-AllWeather},~\cite{Rain-Rcp}, which can learn weather-invariant features. In the deraining context, the colored-image intensity is expressed as

\begin{equation}
    \tilde{\mathcal{C}}(x)=\tau\rho_{w}(x)\mathbf{L}\sigma+(T-\tau)\mathbf{B}\pi
\end{equation}
where $T$ denotes the exposure time and the elapsed time while the weather is passing through a pixel $x$ is $\tau$. $\rho_{w}$ denotes the composition of refraction, specular reflection, and internal reflection. $\mathbf{L}=[{L_r, L_g,L_b}]^{\mathrm{T}}$ with $\mathrm{L}=L_r+L_g+L_b$ and $\mathbf{B}=[{B_r, B_g, B_b}]^{\mathrm{T}}$ with $\mathrm{B}=B_r+B_g+B_b$ denote the color vector of the light brightness and background reflection, respectively. $\sigma=\mathbf{L}/\mathrm{L}$ and $\pi=\mathbf{B}/\mathrm{B}$ represent the light chromaticity of light brightness $\mathbf{L}$ and background reflection $\mathbf{B}$, respectively. To generate the residue channel free from any degradation, one can estimate $\sigma$ via a color constancy algorithm and then apply the normalization as
\begin{equation} 
\mathcal{C}(x)=\frac{\tilde{\mathcal{C}}(x)}{\sigma}=\mathcal{C}_{w}(x)\mathbf{i}+\mathcal{C}_{bg}(x)
\end{equation}
where $\mathbf{i}=(1, 1, 1)^{\mathrm{T}}$, $\mathcal{O}_{w}=\tau\rho_{w}\mathbf{L}$, and $\mathcal{O}_{bg}=(T-\tau)\mathbf{B}/\sigma$. Notably, the light chromaticity and color effect of the spectral sensitivities are canceled during the normalization process. The achromatic nature of weather artifacts implies minimal inter-channel variance, with RGB values converging toward identity in affected regions. Therefore, the RCP can be defined by subtracting the minimum channel of the images from the maximum channel, \emph{i.e.} residue channel, which can eliminate the weather influences to extract object structures.
In this work, considering the foundation of typical adverse weather in the physics of atmospheric scattering, we generalize RCP to all weather types in UAWR. Therefore, we can extract the RCP $\mathbf{R}(x)$ of the initial restoration $\tilde{\mathbf{I}}_{HQ}$ by

\begin{equation}
    \mathbf{R}(x) = \tilde{\mathbf{I}}^M_{HQ}(x)-\tilde{\mathbf{I}}^m_{HQ}(x)
\end{equation}
where $\mathbf{I}^M(x)=\mathrm{max}(\mathbf{I}_r(x), \mathbf{I}_g(x), \mathbf{I}_b(x))$ and $\mathbf{I}^m(x)=\mathrm{mim}(\mathbf{I}_r(x), \mathbf{I}_g(x), \mathbf{I}_b(x))$. Fig.~\ref{fig:vis_lq_iter0_rcp} shows that RCP can provide degradation-free structural contextualized information. 
 To take full advantage of the residual prior, in the 2nd cyclic prompt-based restoration, it complements with $\hat{\mathbf{P}}_c$ in the prompt block (see Fig.~\ref{fig:transformer}) to enrich the texture clues and structural appearance. By combining C2P and EPM, we can form a distinctive cyclic prompt learning framework, providing holistic representations such that the restoration process can be effectively adapted for adverse weather.




\subsection{Prompt Block}
\label{sec:rpm}
We use the transformer architecture in Restormer~\cite{Restormer} as the backbone, which adopts a 4-level transformer encoder and decoder, where each level contains several transformer blocks, illustrated in Fig.~\ref{fig:transformer}. The transformer encoder gradually downsamples the input image $\mathbf{I}_{LQ}\in\mathbb{R}^{H\times W\times C}$ while increasing the channel number, resulting in deep image features $\mathbf{F}_e\in\mathbb{R}^{\frac{H}{8}\times \frac{W}{8}\times 8C}$. The prompt block is inserted before each level of the decoder. As shown in Fig.~\ref{fig:transformer}(a), each block consists of two branches, with the branch selection controlled by the restoration process: the 1st iteration with initial C2P (top branch) and the 2nd with cyclic C2P (bottom branch). 

Specifically, let $\mathbf{X}$ denote the incoming feature fed into a certain prompt block. For the 1st iteration, the C2P $\mathbf{P}_c$ is projected into key and value spaces via two respectively linear layers $\mathbf{W}_k$ and $\mathbf{W}_v$, getting $\mathbf{K}\in\mathbb{R}^{N\times C}=\mathbf{W}_k\mathbf{P}_c$ and $\mathbf{V}\in\mathbb{R}^{N\times C}=\mathbf{W}_v\mathbf{P}_c$, where $\mathbf{X}$ is reshaped to the same size of $N\times C$ and then projected to the query space $\mathbf{V}\in\mathbb{R}^{N\times C}=\mathbf{W}_q\mathbf{X}$. We interact $\mathbf{P}_c$ with $\mathbf{X}$ through a cross-attention mechanism~\cite{SD}
\begin{equation}
    \hat{\mathbf{X}}=\mathcal{A}((\mathbf{Q}, \mathbf{\mathbf{K}}), \mathbf{V})={\rm{Softmax}}(\frac{\mathbf{Q}\mathbf{K}^\mathrm{T}}{\sqrt{d}})\cdot \mathbf{V},
\end{equation}
where $\mathcal{A}$ denotes the cross attention operation and  $\hat{\mathbf{X}}\in\mathbb{R}^{N\times C}$ is the resulted output which is then transformed to $H\times W\times C$.

When the current iteration is the second,  the cyclic prompt $\hat{\mathbf{P}}_c$ generated by EPM and the residual prior $\mathbf{R}$ pass through the bottom branch sequentially. In addition to the cross attention between $\mathbf{X}$ and $\hat{\mathbf{P}}_c$ as the 1st iteration, there is a residual prior modulator (RPM) in the prompt block that conducts conditional modulation based on $\mathbf{R}$. As shown in Fig.~\ref{fig:transformer}(b), two $3\times 3$ convolutions followed by two cross-hierarchy information mining block (CHIMB)~\cite{CVHSSR} are leveraged to learn informative residual feature $\mathbf{F}_r$ from  $\mathbf{R}$ with large kernel attention and channel attention. Next, another two $3\times 3$ convolutional layers with two respective $1\times 1$ convolution layers are applied to generate a pair of affine transformation parameters ($\alpha$, $\beta$). Therefore, the residual prior can modulate the attentive features $\tilde{\mathbf{X}}$ as follows:
\begin{equation}
\hat{\mathbf{X}}=\alpha\odot \tilde{\mathbf{X}} + \beta,
\end{equation}
where $\hat{\mathbf{X}}$ denotes the modulated feature and $\odot$ denotes the element-wise multiplication.

Therefore, the overall cyclic prompt learning of both iterations can be defined as 
\begin{equation}
    \hat{\mathbf{X}}= \begin{cases}
   \mathcal{A}(\mathbf{P}_c, \mathbf{X}), & iter.=1, \\
    \mathcal{R}( \mathbf{R}, \mathcal{A}(\hat{\mathbf{P}}_c, \mathbf{X} ) ), & iter.=2,
    \end{cases}
\end{equation}
where $\mathcal{R}$ represents the composite function of RPM. As the prompt blocks are deployed from low to high levels, one can dynamically integrate the prompt components into multi-scale features to provide sufficient guidance.

\subsection{Theoretical Analysis of Cyclic Prompt}

In this work, the restoration network $f_{\theta}$ is conditioned on a prompt $\mathbf{P}$ that encodes prior knowledge. Therefore, our CyclicPrompt framework can be viewed as performing two consecutive optimizations:

\begin{equation}
\begin{aligned}
\theta_1&=\mathop{\arg\min}_{\theta}\mathcal{L}_1(\theta)\triangleq\Vert{f_{\theta}(\mathbf{I}_{LQ};\mathbf{P}_c)-\mathbf{I}_{HQ}}\Vert_1\\
\theta_2&=\mathop{\arg\min}_{\theta}\mathcal{L}_2(\theta)\triangleq\Vert{f_{\theta}(\mathbf{I}_{LQ};\hat{\mathbf{P}}_c, \mathbf{R})-\mathbf{I}_{HQ}}\Vert_1
\end{aligned}
\end{equation}

In fact, this is an iterative refinement with a better-conditioned prompt. Here, $\theta_1$ and $\theta_2$ denote the weight sets in two iterations. $\mathcal{L}_1$ and $\mathcal{L}_2$ corresponds to the two terms in Eq.~(1). Intuitively, the prompt $\hat{\mathbf{P}}_c$ is closer to the true latent clean image manifold than $\mathbf{P}_c$. This is because EPM explicitly removes weather-specific information and replaces it with the weather-free textual prompt revealing the clean scene context and residual prior providing degradation-invariant structural cues (see Fig.~\ref{fig:vis_lq_iter0_rcp}). Thus, the 2nd iteration is conditioned on a more accurate prior that reduces the sensitivity to the degradation interference in LQ images, which can enhance the generalizability.

Besides, in our method, the 1st iteration can be seen to constrain the solution space $\mathcal{S}$ conditioned on the C2P $\mathbf{P}_c$ and the 2nd iteration with the  EPM project $\mathcal{S}$ onto the manifold $\mathcal{M}$ of the clean image $\mathbf{I}_{HQ}$. Therefore, the cyclic process minimizes the manifold discrepancy $\Delta_{\mathcal{M}}$ by

\begin{equation}
\Delta_{\mathcal{M}}=\mathrm{dist}(\mathcal{S}(\mathbf{P}_c), \mathcal{M})-\mathrm{dist}(\mathcal{S}(\hat{\mathbf{P}}_c, \mathbf{R}), \mathcal{M})>0
\end{equation}

In this way, the C2P and EPM C2P create complementary effects and improve the optimization landscape.

\section{Experiments}
\label{sec:experiments}
We conducted experiments to evaluate the effectiveness of the proposed method. This section begins by outlining the experimental settings, followed by a detailed presentation of the qualitative and quantitative results. Lastly, we perform ablation studies to analyze the contributions of key components.

\begin{table*}[t]
    \caption{Quantitative Comparisons in Terms of PSNR and SSIM with State-of-the-Art Adverse Weather Removal Methods. * Denotes the Model Retrained on the All-Weather Dataset. \textbf{Text} Denotes the Best Performance and \underline{Underline} Denotes the Second Best.}
    \begin{center}
    \begin{tabular}{c c c c c c c c c c c c }
    \hline
    \hline
    \multirow{2}{*}{\textbf{Types}} & \multirow{2}{*}{\textbf{Methods}} & \multicolumn{2}{c}{\textbf{Outdoor-Rain}~\cite{HRGAN}} & \multicolumn{2}{c}{\textbf{RainDrop}~\cite{Dataset-Raindrop}} & \multicolumn{2}{c}{\textbf{Snow100K-S}~\cite{Dataset-Snow100K}} & \multicolumn{2}{c}{\textbf{Snow100K-L}~\cite{Dataset-Snow100K}} & \multicolumn{2}{c}{\textit{\textbf{Average}}} \\
    \cmidrule(r){3-4} \cmidrule(r){5-6} \cmidrule(r){7-8} \cmidrule(r){9-10}
    \cmidrule(r){11-12} & & PSNR & SSIM & PSNR & SSIM & PSNR & SSIM & PSNR & SSIM & PSNR & SSIM \\
	\midrule[0.5pt]
    \multirow{4}{*}{Rain+Fog} 
    & Pix2Pix\textsubscript{[CVPR'17]}~\cite{Pix2Pix} & 19.09 & 0.7100 & - & - & - & - & - & - & - & -\\
    & HRGAN\textsubscript{[CVPR'19]}~\cite{HRGAN} & 21.56 & 0.8550 & - & - & - & - & - & - & - & -\\
    & PCNet\textsubscript{[TIP'21]}~\cite{PCNet} & 26.19 & 0.9015 & - & - & - & - & - & - & - & -\\
    & MPRNet\textsubscript{[CVPR'21]}~\cite{MPRNet} & 30.44 &	0.9235 & 19.92 & 0.8385 & 22.13 & 0.7824 & 20.25 & 0.6628 & 23.19 & 0.8018 \\ 
    \midrule[0.5pt]			
    \multirow{4}{*}{Raindrop}
    & AttentiveGAN\textsubscript{[CVPR'18]}~\cite{Dataset-Raindrop} & - & - & 31.59 & 0.9170 & - & - & - & - & - & - \\
    & RaindropAttn\textsubscript{[ICCV'19]}~\cite{quan2019deep} & - & - & 31.44 & 0.9263 & - & - & - & - & - & - \\
    & CCN\textsubscript{[CVPR'21]}~\cite{Dataset-RainDS} & - & - & 31.34 & 0.9286 & - & - & - & - & - & - \\
    & IDT\textsubscript{[PAMI'22]}~\cite{RainTransformer-IDT} & 14.79 & 0.5416  & 31.87	& 0.9313 & 25.36 & 0.8560 & 21.52 & 0.7565 & 24.01 & 0.7763  \\ 
    \midrule[0.5pt]
    \multirow{4}{*}{Snow}
    & DesnowNet\textsubscript{[TIP'18]}~\cite{Dataset-Snow100K} & - & - & - & - & 32.33 & 0.9500 & 27.17 & 0.8983 & - & - \\
    & JSTASR\textsubscript{[ECCV'20]}~\cite{JSTASR} & - & - & - & - & 31.40 & 0.9012 & 25.32 & 0.8076 &  - & - \\
    & DDMSNet\textsubscript{[TIP'21]}~\cite{DDMSNet} & - & - & - & - & 34.34 & 0.9445 & 28.85 & 0.8772 & - & - \\
    & MPRNet\textsubscript{[CVPR'21]}~\cite{MPRNet} & - & - & - & - & 34.97 & 0.9457 & 29.76 & 0.8949 & - & - \\ 

    \hline
    \hline
    \multirow{3}{*}{General} 

     & MPRNet*\textsubscript{[CVPR'21]}~\cite{MPRNet}  & 29.41 & 0.8990 & 30.76 & 0.9100 & - & - & 29.25 & 0.8803 & - & - \\
     & Restormer*\textsubscript{[CVPR'22]}~\cite{Restormer} & 28.54 & 0.8856 & 30.61 & 0.9027 & - & - & 29.09 & 0.8725  & - & - \\ [1pt]   
     & Uformer*\textsubscript{[CVPR'22]}~\cite{Uformer} & 26.72 & 0.8453 & 28.69 & 0.8819 & - & - & 27.92 & 0.8476  & - & - \\ [1pt] 

    \hline
    \hline
    \multirow{13}{*}{All-in-One} 
    & All-in-one\textsubscript{[CVPR'20]}~\cite{Dataset-AllWeather} & 24.71 & 0.8980 & 31.12 & 0.9268 & 28.33 & 0.8820 & 28.05 & 0.9023 & - & - \\ [1pt] 
    & TransWeather\textsubscript{[CVPR'22]}~\cite{TransWeather} & 28.83 & 0.9000 & 30.17 & 0.9157	& 32.51 & 0.9341 & 29.31 & 0.8879 & 30.21 & 0.9094   \\ [1pt] 
    & TKL\&MR\textsubscript{[CVPR'22]}~\cite{TUM_chen} & 29.92 & 0.9167 & 30.99 & 0.9274	& 34.80 & 0.9483 & 30.24 & 0.9020 & 31.49 & 0.9236  \\ [1pt] 
    & WeatherDiff$_{64}$\textsubscript{[PAMI'23]}~\cite{WeatherDiff} & 29.64 & 0.9312 & 30.71 & 0.9312 & 35.83 & \textbf{0.9566} & 30.09 & 0.9041 & 31.57 & 0.9308  \\
    & WeatherDiff$_{128}$\textsubscript{[PAMI'23]}~\cite{WeatherDiff} & 29.72 & 0.9216 & 29.66 & 0.9225 & 35.02 & 0.9516 & 29.58 & 0.8941 & 31.00 & 0.9225 \\
    & AWRCP\textsubscript{[ICCV'23]}~\cite{AWRCP} & 31.39 & 0.9329 & 31.93 & 0.9314	& 36.92 & 0.9652 & 31.92 & \textbf{0.9341} & 33.04 & 0.9409  \\
    & Tan \textit{et al.}\textsubscript{[TIP'24]}~\cite{tan_tip24}& 30.00 & 0.9152 & 31.72 & 0.9325 & 29.85 & 0.8928 & - & - & - & - \\ 
    & Art\textsubscript{[MM'24]} ~\cite{Art} & 29.81 & 0.9088 & 31.54 & 0.9338 & -	& -	& 30.61 & 0.9083 & - & - \\
    & T$^3$-DiffWeather\textsubscript{[ECCV'24]}~\cite{T3-DiffWeather} & \underline{32.52} & \underline{0.9339} & \textbf{32.70} & 0.9414 & \textbf{37.55} & 0.9641 & 31.11 & 0.9180 & 33.45 & 0.9391 \\
    & GridFormer\textsubscript{[IJCV'24]}~\cite{wang2024gridformer} & 31.87 & 0.9335 & 32.39 & 0.9362 & 37.46 & 0.9640 & 31.71 & 0.9231 & 33.36 & 0.9392 \\
    & \textbf{CyclicPrompt (ours)} & \textbf{32.81} & \textbf{0.9371} & \underline{32.57} & \textbf{0.9454}  & \underline{37.50} & \underline{0.9655} & \textbf{32.16} & \underline{0.9265} & \textbf{33.76} & \textbf{0.9436} \\
    
    & \textbf{CyclicPrompt} (w/o text) &	32.33 	& 0.9333  & 32.30 & \underline{0.9453} & 37.34 & 0.9649 & \underline{32.02} & 0.9254 & 33.50 & \underline{0.9422} \\ 
        
      &  \textbf{CyclicPrompt} (w/ random text) & 32.28 & 0.9332 & 32.56 & 0.9450 & 37.33 & 0.9649 	& 32.01 & 0.9251 & \underline{33.55} & 0.9421 \\
        \bottomrule

    \hline
    \hline 
    \end{tabular}
    \end{center}
\label{tab:psnr_ssim}
\end{table*}

\subsection{Implementation Details}
Following the transformer setting in~\cite{Restormer}, in CyclicPrompt, from level 1 to level 4 of the encoder, the number of transformer blocks [$B_1$, $B_2$, $B_3$, $B_4$] is set to [4, 6, 6, 8] and the channel numbers are [48, 96, 192, 384]. Since the decoder employs a two-iteration design, the number of transformer blocks is reduced by half with the same channel numbers as the encoder for computational efficiency. The prompt block is incorporated before each decoder level, totaling 3 blocks in the overall CyclicPrompt network.

We train our model with AdamW optimizer (\(\beta_1=0.9, \beta_2=0.9999 \), weight decay=1e-4 ) with a batch size of 4 for 800K iterations. The initial learning rate is 2e-4 and gradually reduces to 1e-6 by cosine annealing strategy. The input images are randomly cropped into $256\times 256$ during training. Randomly vertical and horizontal flips are used for data augmentation. All the models are trained on NVIDIA RTX 4090 GPUs.

\subsection{Dataset}
We conduct experiments on the All-Weather dataset~\cite{Dataset-AllWeather} mainly degraded by four typical adverse weather conditions: \textbf{Rain+Fog}, \textbf{Raindrop}, and \textbf{Snow}. This dataset includes 18,069 images as the training data, comprising 9,001 synthesized snowy images from Snow100K~\cite{Dataset-Snow100K}, 818 images with raindrops on windows or lenses from RainDrop~\cite{Dataset-Raindrop}, and 8,250 heavy rain images from Outdoor-Rain~\cite{HRGAN} for Rain+Fog. Notably, the Outdoor-Rain subset captures both rain streaks and fog-like degradation caused by rain accumulation. Due to the imbalance in the number of training samples for different weather conditions, particularly the limited size of the RainDrop subset, we perform a 10-fold resampling of the Raindrop images to ensure balanced training. 

For testing, there are also 750 images from Outdoor-Rain Test1~\cite{HRGAN} (Rain+Fog), 58 images from RainDrop TestA~\cite{Dataset-Raindrop}, 16,611 images from Snow100K-S~\cite{Dataset-Snow100K}, and 16,801 images from Snow100K-L~\cite{Dataset-Snow100K}. We utilize peak signal-to-noise ratio (PSNR) and structural similarity (SSIM) as the metrics to evaluate the quantitative performance of our method.

\subsection{Comparison with State-of-the-art Methods}
We compare the proposed CyclicPrompt with recent state-of-the-art (SotA) adverse weather removal methods, which can be categorized into 3 groups: 
\begin{itemize}
    \item \textbf{Task-specific methods}: Pix2Pix~\cite{Pix2Pix}, HRGAN~\cite{HRGAN}, PCNet~\cite{PCNet}, and MPRNet~\cite{MPRNet} for rain+fog; AttentiveGAN~\cite{Dataset-Raindrop}, RaindropAttn~\cite{quan2019deep}, CCN~\cite{Dataset-RainDS}, and IDT~\cite{RainTransformer-IDT} for raindrop; DesnowNet~\cite{Dataset-Snow100K}, JSTASR~\cite{JSTASR}, DDMSNet~\cite{DDMSNet}, and MPRNet~\cite{MPRNet} for snow.

    \item \textbf{General Image Restoration Methods retrained on the All-Weather dataset}: MPRNet*~\cite{MPRNet}, Restormer*~\cite{Restormer}, and Uformer*~\cite{Uformer}.

    \item \textbf{UAWR methods}: All-in-one~\cite{Dataset-AllWeather}, TransWeather~\cite{TransWeather}, TKL\&MR~\cite{TUM_chen}, WeatherDiff~\cite{WeatherDiff}, AWRCP~\cite{AWRCP}, Tan \textit{et al}.~\cite{tan_tip24}, Art~\cite{Art}, T$^3$-DiffWeather~\cite{T3-DiffWeather}, and GridFormer~\cite{wang2024gridformer}
\end{itemize}

\begin{figure*}[!t]
	\centering
	\includegraphics[width=1\linewidth]{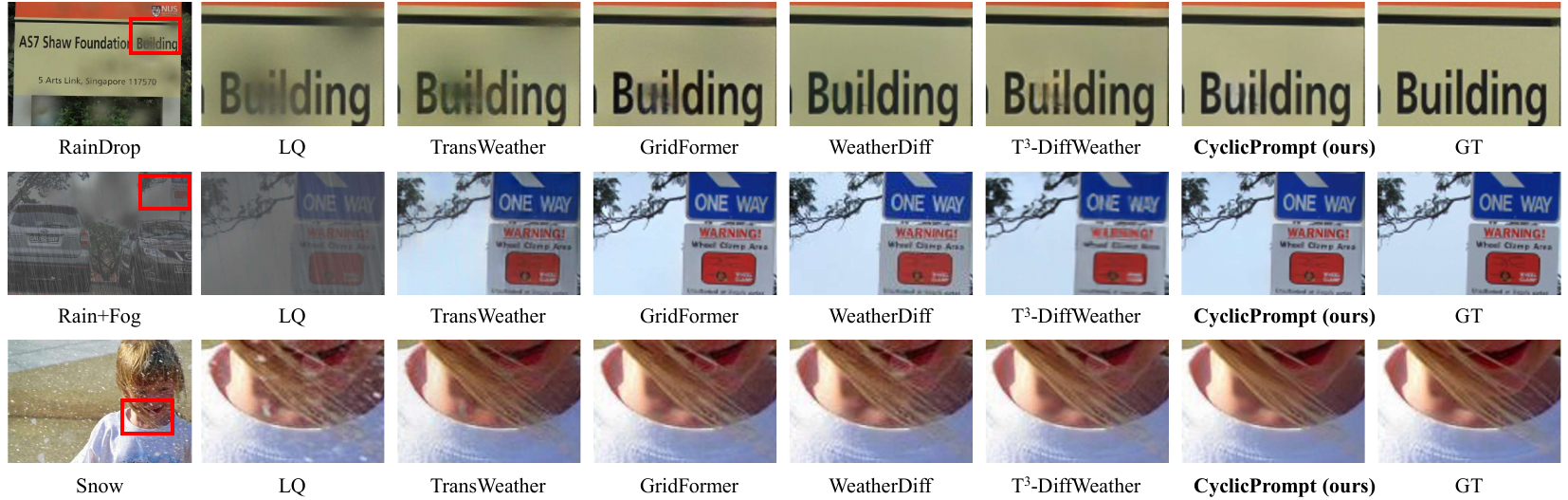}
	\caption{Visual Comparisons with different UAWR methods, \emph{i.e.} TransWeather~\cite{TransWeather}, GrideFormer~\cite{wang2024gridformer}, T$^3$-DiffWeather~\cite{T3-DiffWeather}, WeatherDiff~\cite{WeatherDiff} and our CyclicPrompt on the RainDrop~\cite{Dataset-Raindrop}, Outdoor-Rain~\cite{HRGAN} and Snow100K-L~\cite{Dataset-Snow100K} datasets.}
	\label{fig:allweather_vis}
\end{figure*}

\begin{table*}[t]
    \caption{Quantitative Results under Real-World Scenarios. \textbf{Text} Denotes the Best Performance and \underline{Underline} Denotes the Second Best. The FLOPs are computed on the input image of size $256\times 256$. We calculate the average FPS on the RainDS ($1296\times 768$) and SPA ($512\times 512$) datasets.}
    \centering
    \resizebox{\linewidth}{!}{
    \begin{tabular}{cccccccccccc c }
        \toprule
        \multirow{2}{*}{Methods}& \multicolumn{3}{c}{\textbf{SPA}~\cite{Dataset-SPA}} & \multicolumn{3}{c}{\textbf{RainDS}~\cite{Dataset-RainDS}} & \multicolumn{3}{c}{\textbf{Snow100K-real}~\cite{Dataset-Snow100K}} & Params. & GFlops  & FPS \\ 
        \cmidrule(r){2-4} \cmidrule(r){5-7} \cmidrule(r){8-10} 
        ~ & PSNR & SSIM & MUSIQ$\uparrow$ & PSNR & SSIM & MUSIQ$\uparrow$ & MUSIQ$\uparrow$ & MANIQA$\uparrow$ & NIQE$\downarrow$  & (M) & (G) & (frames/s) \\
        \midrule
        TransWeather~\cite{TransWeather} & 27.96 & 0.8677 & 45.83 & 21.78 & 0.690 & 49.64 & 60.56 & 0.388 & 3.042 & 37.7 & 6.13 & 56.7784 \\

        WeatherDiff$_{128}$~\cite{WeatherDiff} & 29.26 & 0.9279 & \underline{46.32} & 22.40  & 0.6405 & \textbf{59.00} & \textbf{62.48} & \underline{0.396} & 2.970 & 85.6 & 65.9$\times$169$\times$25 & 0.0095 \\
        T$^3$-DiffWeather~\cite{T3-DiffWeather} & 31.36 & 0.9340 & 45.01 & \underline{23.12} & \underline{0.6461} & 45.01 & 61.65 & \textbf{0.397} & \underline{2.910} & 69.4 & 59.8$\times$2 & 3.0385 \\
        T$^3$-DiffWeather (4 steps)~\cite{T3-DiffWeather} & \underline{31.65} & \textbf{0.9367} & 46.10 & \textbf{23.17} & 0.6388 & 43.58 & 61.70 & \textbf{0.397} & 2.922 & 69.4 & 59.8$\times$4 &  1.7252 \\
        GridFormer~\cite{wang2024gridformer} & 31.18 & 0.9269 & 46.23 & 22.89 & 0.6377 & 46.85 & 61.49 & 0.395 & 2.912 & \underline{30.1} & 251.35 & 2.1546 \\

        \textbf{CyclicPrompt (ours)} & \textbf{31.95} & \underline{0.9347} & \textbf{46.51} & 23.10 & \textbf{0.6620} & \underline{50.32} & \underline{61.72} & 0.395 & \textbf{2.893} & \textbf{29.9} & 230.42 & 2.5231 \\
        \bottomrule
    \end{tabular}
    }
    \label{tab:realworld}
\end{table*}

For fair comparisons, we evaluate the results of the methods using their official codes or pre-trained weights once provided. Otherwise, we directly cite the reported results in their papers.

\textbf{Quantitative Comparison}.
Table~\ref{tab:psnr_ssim} provides the quantitative results of all methods across three adverse weather types. We can see that the task-specific methods, \emph{e.g.}, MPRNet~\cite{MPRNet} and IDT~\cite{RainTransformer-IDT}, produce promising results on the weather type when applying their corresponding customized models or even better than several UAWR methods (\emph{e.g.}, All-in-one~\cite{Dataset-AllWeather}). However, as shown in Table~\ref{tab:psnr_ssim}, they fail to be generalized to other out-of-distribution weather types, such as MPRNet on raindrops and IDT on snow, where their PSNR and SSIM decrease significantly. When we retrain the general restoration methods on the same All-Weather dataset according to the UAWR setting, one can observe that the resultant models show better generalizability on three different datasets (MPRNet \emph{v.s.} MPRNet*). Our CyclicPrompt outperforms existing methods by a significant margin across almost all weather types and achieves the best on average. Compared to the recent most SotA method T$^3$-DiffWeather, our method exceeds it by over 0.3dB in average PSNR. This is mainly due to our holistic prompt construction and distinctive cyclic prompt learning design, which is preferable for UAWR.

Besides, we also implement two model variants by: 1) removing the textual prompt (w/o text), thus the model relies solely on visual prompts and learnable conditional vectors; 2) randomly shuffling the paired textual descriptions in the test set (w/ random text), feeding non-corresponding captions as textual prompts to the model. In Table~\ref{tab:psnr_ssim}, we can see that removing textual prompts causes only a slight performance drop, indicating that the model effectively leverages visual and learnable prompts to maintain restoration quality. Though mismatched textual prompts lead to moderate degradation, it still performs competitively. Moreover, all our model variants achieve comparable performance across most datasets and superior performance on average against existing methods, demonstrating the robustness of our method in practical applications.

\textbf{Qualitative Comparison}.
In Fig.~\ref{fig:allweather_vis}, we visualize the restoration results of all compared methods. It can be observed that, under various weather conditions, our method can consistently remove the adverse effects of unfavorable weather from corrupted images and preserve plausible textures. Especially, in the rain+fog degradation, our method effectively recovers obscured scenes with more fine-grained details.

\begin{figure*}[!t]
	\centering
	\includegraphics[width=1\linewidth]{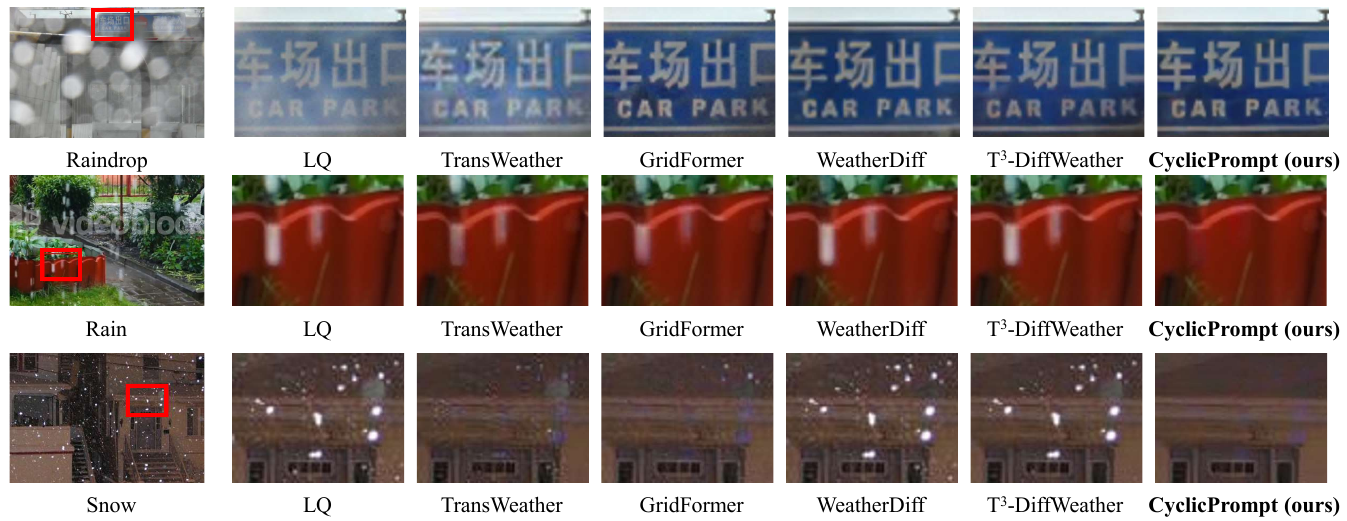}
	\caption{Qualitative comparison of different adverse weather removal methods, \emph{i.e.}, TransWeather~\cite{TransWeather}, GridFormer~\cite{wang2024gridformer}, WeatherDiff~\textit{et al}.~\cite{WeatherDiff}, T$^3$-DiffWeather~\cite{T3-DiffWeather}, and our CyclicPrompt under real scenarios. Test images are from RainDS\cite{Dataset-RainDS}, SPA-Data\cite{Dataset-SPA} and Snow100K-real\cite{Dataset-Snow100K}.}  
	\label{fig:real_vis}
\end{figure*}

\begin{table*}[t]
\caption{Quantitative Comparisons on Five Tasks. \textbf{Text} and Denote the Best and Performance and \underline{Underline} Denotes the Second Best.}
\resizebox{\textwidth}{!}{
\begin{tabular}{ccccccccccccc}
\toprule
\multicolumn{1}{c}{\multirow{2}{*}{\textbf{Methods}}} & \multicolumn{2}{c}{\begin{tabular}[c]{@{}c@{}}\textbf{Deraining}\\ Rain100L~\cite{Dataset-Rain100L}\end{tabular}} & 
\multicolumn{2}{c}{\begin{tabular}[c]{@{}c@{}}\textbf{Dehazing}\\ SOTS~\cite{Dataset-RESIDE-OTS}\end{tabular}} & 
\multicolumn{2}{c}{\begin{tabular}[c]{@{}c@{}}\textbf{Denoising}\\ BSD68~\cite{Dataset-BSD400}\end{tabular}} & 
\multicolumn{2}{c}{\begin{tabular}[c]{@{}c@{}}\textbf{Deblurring}\\ GoPro~\cite{Dataset-GOPRO}\end{tabular}} & 
\multicolumn{2}{c}{\begin{tabular}[c]{@{}c@{}}\textbf{Enhancement}\\ LOL~\cite{Dataset-LOL}\end{tabular}} & \multicolumn{2}{c}{\textbf{\textit{Average}}} \\ \cmidrule(r){2-3} \cmidrule(r){4-5} \cmidrule(r){6-7} \cmidrule(r){8-9} \cmidrule(r){10-11} \cmidrule(r){12-13}
\multicolumn{1}{c}{} & PSNR & SSIM & PSNR & SSIM  & PSNR & SSIM & PSNR & SSIM & PSNR & SSIM & PSNR & SSIM \\
\midrule
HINet\textsubscript{[CVPR'21]}\cite{HINet} & 35.67 & 0.969 & 24.74 & 0.937 & 31.00 & 0.881 & 26.12 & 0.788 & 19.47 & 0.800 & 27.40 & 0.875 \\
DGUNet\textsubscript{[CVPR'22]}\cite{DGUNet} & 36.62 & 0.971 & 24.78 & 0.940 & 31.10 & 0.883 & 27.25 & 0.837 & 21.87 & 0.823 & 28.32 & 0.891 \\
MIRNetV2\textsubscript{[CVPR'22]}\cite{MIRNetV2} & 33.89 & 0.954 & 24.03 & 0.927 & 30.97 & 0.881 & 26.30 & 0.799 & 21.52 & 0.815 & 27.34 & 0.875 \\
SwinIR\textsubscript{[ICCV'21]}\cite{SwinIR} & 30.78 & 0.923 & 21.50 & 0.891 & 30.59 & 0.868 & 24.52 & 0.773 & 17.81 & 0.723 & 25.04 & 0.835 \\
Restormer\textsubscript{[CVPR'22]}\cite{Restormer} & 34.81 & 0.962 & 24.09 & 0.927 & 31.49 & 0.884 & 27.22 & 0.829 & 20.41 & 0.806 & 27.60 & 0.881 \\
NAFNet\textsubscript{[ECCV'22]}\cite{NAFNet} & 35.56 & 0.967 & 25.23 & 0.939 & 31.02 & 0.883 & 26.53 & 0.808 & 20.49 & 0.809 & 27.76 & 0.881 \\
\midrule
DL\textsubscript{[TPAMI'19]}\cite{DL} & 21.96 & 0.762 & 20.54 & 0.826 & 23.09 & 0.745 & 19.86 & 0.672 & 19.83 & 0.712 & 21.05 & 0.743 \\
TAPE\textsubscript{[ECCV'22]}\cite{TAPE} & 29.67 & 0.904 & 22.16 & 0.861 & 30.18 & 0.855 & 24.47 & 0.763 & 18.97 & 0.621 & 25.09 & 0.801 \\
AirNet\textsubscript{[CVPR'22]}\cite{AirNet} & 32.98 & 0.951 & 21.04 & 0.884 & 30.91 & 0.882 & 24.35 & 0.781 & 18.18 & 0.735 & 25.49 & 0.846  \\
IDR\textsubscript{[CVPR'23]}\cite{IDR} & 35.63 & 0.965 & 25.24 & 0.943 & \textbf{31.60} & 0.887 & 27.87 & 0.846 & 21.34 & 0.826 & 28.34 & 0.893 \\
InstructIR-5D\textsubscript{[ECCV'24]}\cite{InstructIR} & \underline{36.84} & \underline{0.973} & 27.10 & 0.956 & \underline{31.40} & 0.887 & \underline{29.40} & \underline{0.886} & \textbf{23.00} & \underline{0.836} & \underline{29.55} & \underline{0.907} \\
DA-CLIP\textsubscript{[ICLR'24]}\cite{DA-CLIP} & - & - & 31.98 & 0.971 & - & - & 27.03 & 0.810 & \underline{22.09} & 0.810 & - & - \\
PromptIR\textsubscript{[NeurIPs'24]}\cite{PromptIR} & 36.37	& 0.972 & 30.58	& \underline{0.974} & 31.31 & 0.880 & - & - & - & - & - & -\\
NDR\textsubscript{[TIP'24]}\cite{NDR} & 35.42 & 0.969 & 28.64 & 0.962 & 31.36 & 0.887 & - & - & - & - & - & - \\ 
DiffUIR\textsubscript{[CVPR'24]}\cite{zheng2024selective} & 30.35 & 0.920 & 31.08	& 0.935 & - & - & 27.69 &	0.845 & 21.93 & 	0.826 & - & - \\ 

\midrule

\textbf{CyclicPrompt (ours)} & \textbf{37.03} & \textbf{0.989} & \textbf{33.02} & \textbf{0.983} & 31.33 & \textbf{0.941} & \textbf{29.42} & \textbf{0.925} &	21.79 & \textbf{0.837} & \textbf{30.52} & \textbf{0.935} \\

\bottomrule
\end{tabular}
}
\label{tab:metrics_5d}
\end{table*}

\textbf{Real Scenario Comparison}. We also evaluate the performance of our CyclicPrompt in real-world scenarios under different weather conditions and compare it with more recent methods. We use the SPA~\cite{Dataset-SPA}, RainDS~\cite{Dataset-RainDS}, and Snow100K-real~\cite{Dataset-Snow100K} datasets as the testing sets. We measure the performance using PSNR, SSIM, and a non-reference metric MUSIQ~\cite{ke2021musiq}. For Snow100K-real, since the ground truth images are unavailable, we apply the non-reference metrics MUSIQ, MANIIQA~\cite{yang2022maniqa}, and NIQE~\cite{mittal2012niqe} for quantitative evaluation. We also measure the model complexity by calculating the model parameters (Params., M), computation costs (GFLOPs, G), and average inference speed in FPS on two datasets: RainDS ($1296\times768$) and SPA dataset ($512\times512$). 

As illustrated in Table~\ref{tab:realworld}, Transweather requires the fewest computations with the fastest speed but more parameters than CyclicPrompt and suffers from significant performance reduction. T$^3$-DiffWeather performs efficiently but with over $2\times$ parameters and slightly lower scores than CyclicPrompt. To make a more intuitive comparison, we increase the iteration step of T$^3$-DiffWeather to 4, ensuring it has similar computational costs (239.2G) to CyclicPrompt. The results reveal that the model does not yield apparent improvements, which still underperforms CyclicPrompt across most metrics. We can also see that its efficiency is slowed down, which shows fewer FPS than ours. GridFormer shows relatively heavier model complexity but is not adept in real-world weather, which performs worse and more slowly than our method. Though WeatherDiff$_{128}$ is comparable to ours, this method involves about $2.5\times$ more parameters and $1200\times$ more FLOPs. CyclicPrompt achieves competitive performance with obviously smaller model complexity and over $200\times$ FPS.
\\ \indent In Fig.~\ref{fig:real_vis}, we illustrate the visual comparisons on real-world datasets. It is evident that both WeatherDiff$_{128}$ and T$^3$-DiffWeather are unable to suppress the weather interferences, while CyclicPrompt can produce more visually appealing images. Such quantitative and qualitative performance showcases the superior trade-off between the adaptability, efficiency, and generalizability of CyclicPrompt in practical applications. \\
\indent\textbf{Generalizing to More Diverse Degradations}.
To assess the further effectiveness and generalizability of CyclicPrompt, we apply our method to five restoration tasks: deraining, dehazing, motion deblurring, low-light enhancement, and denoising as in~\cite{IDR}. Table~\ref{tab:metrics_5d} shows the quantitative performance on Rain100L~\cite{Dataset-Rain100L} for deraining, BSD68~\cite{Dataset-BSD400} for denoising, SOTS~\cite{Dataset-RESIDE-OTS} for dehazing, GoPro~\cite{Dataset-GOPRO} for deblurring, and LOL~\cite{Dataset-LOL} for low-light enhancement. Here, we compare CyclicPrompt with recent state-of-the-art general restoration methods: HINet~\cite{HINet}, DGUNet~\cite{DGUNet}, MIRNetV2~\cite{MIRNetV2}, SwinIR~\cite{SwinIR}, Restormer~\cite{Restormer}, NAFNet~\cite{NAFNet} and all-in-one restoration methods: DL~\cite{DL}, TAPE~\cite{TAPE}, AirNet~\cite{AirNet}, IDR~\cite{IDR}, InstructIR-5D~\cite{InstructIR}, DA-CLIP~\cite{DA-CLIP}, PromptIR~\cite{PromptIR}, NDR~\cite{NDR}, and DiffUIR~\cite{zheng2024selective}. As illustrated, our CyclicPrompt achieves impressive performance across almost all the datasets in PSNR and SSIM, emphasizing its excellent superiority to diverse degradations.

\subsection{Ablation Study}
\begin{figure}[] 
    \centering  
    \includegraphics[width=\linewidth]{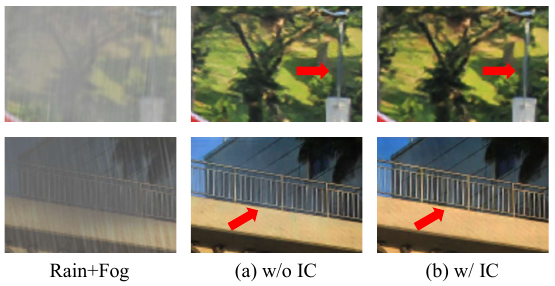}  
    \caption{Visual comparisons between the model without and with the input-conditional vectors, where ``IC'' denotes the input-conditional vector.}
    \label{fig:abl_vis_input_vec}
\end{figure}

In this subsection, we conduct ablation studies to investigate the effects of all the proposed components in CyclicPrompt. All the models are trained for 300K iterations and evaluated on Outdoor-Rain~\cite{HRGAN}, RainDrop~\cite{Dataset-Raindrop}, and Snow100K-L~\cite{Dataset-Snow100K} datasets. We report their average performance for comparison. \\ \indent \textbf{Effect of C2P}. As discussed in Sec.~\ref{sec:c2p}, the initial C2P consists of weather-specific knowledge, learnable input-conditional vector, and textual prompt. We denote the variant without any prompt as the baseline model. As shown in Table~\ref{tab:abl_C2P}, to verify the effectiveness of the initial C2P, we first induce the weather-specific knowledge in the baseline, which exploits the pre-trained CLIP image encoder to extract visual degradation features, prompting the model to improve. The input-conditional vector can be seen as a dynamic prompt for input content, which can calibrate and adapt the weather-dedicated information to complex weather, thereby providing a favorable gain of 0.21dB. It can also be seen that the two components cooperate well and achieve about 0.3dB gains in PSNR, which demonstrates their collaborative potential in restoration capacity improvement. Additionally, in Fig.~\ref{fig:abl_vis_input_vec}, we conduct visual comparisons between the models with and without the input-conditional vectors, where the results validate that the latter can better remove the weather. Interestingly, the textual prompt corresponding to the weather scene is beneficial for restoration as it can incorporate explicit scene contexts and semantics, even if it stems from LQ observations. Hence, the model with full C2P achieves the best.


\begin{table}
    \caption{The Effectiveness of C2P.}
    \centering
    \resizebox{\linewidth}{!}{
    \begin{tabular}{cccc|cc}
        \toprule[1pt]
        \multirow{2}{*}{\textbf{Baseline}} & \textbf{Weather-specific} & \textbf{Input-conditional} & \textbf{Textual} & \multirow{2}{*}{PSNR} & \multirow{2}{*}{SSIM} \\
        ~ & \textbf{Knowledge} & \textbf{Vector} & \textbf{Prompt} & ~ & ~\\
        \midrule[0.5pt]
        \ding{51} & & & & 30.43 & 0.9176  \\
        \ding{51} & \ding{51} & & &  30.47 & 0.9180  \\
        \ding{51} & \ding{51} & \ding{51} & & 30.68 & 0.9195  \\
        \ding{51} & \ding{51} & & \ding{51} & 30.76	& 0.9201  \\ 
        \ding{51} & \ding{51} & \ding{51} & \ding{51} & 
        \textbf{31.07} & \textbf{0.9229}  \\

        \bottomrule[1pt]
    \end{tabular}
    }
    \label{tab:abl_C2P}
\end{table}

\begin{table}[t] 
\renewcommand\arraystretch{}
\caption{The Model Performance across Three Degradation Scenarios under Varying Vector Lengths.}
\label{tab:prefixlen}
\centering
\resizebox{\linewidth}{!}{
\begin{tabular}{c cc cc cc cc }
        \toprule
        \multirow{2}{*}{Vector Number} 
        & \multicolumn{2}{c} {Outdoor-Rain} 
        & \multicolumn{2}{c} {RainDrop} 
        & \multicolumn{2}{c} {Snow100K} 
        & \multicolumn{2}{c} {Average} \\
        \cmidrule(r){2-3} \cmidrule(r){4-5} \cmidrule(r){6-7} \cmidrule(r){8-9} 
        ~ & PSNR & SSIM  & PSNR & SSIM  & PNSR & SSIM & PSNR & SSIM \\
        \midrule
        0	& 31.95	& 0.9384	& 29.87	& 0.9157	& 30.67	& 0.9111	& 30.83	& 0.9217 \\
        4	& 31.87	& 0.9390	& 30.18	& 0.9171	& 30.56	& 0.9119	& 30.87	& 0.9227 \\
        8	& \textbf{32.01}	& \textbf{0.9405}	& \textbf{30.74}	& \textbf{0.9206}	& \textbf{30.80}	& \textbf{0.9142}	& \textbf{31.18}	& \textbf{0.9251} \\ 
        12	& 31.72	& 0.9387	& 29.13	& 0.9167	& 30.27	& 0.9112	& 30.37 & 0.9222 \\
        \bottomrule
    \end{tabular}
}
\end{table}

\begin{table}[t]
    \caption{The Effectiveness of EPM.}
    \centering
    \resizebox{\linewidth}{!}{
    \begin{tabular}{ccc|cc}
        \toprule[1pt]
        \textbf{Baseline+C2P} & \textbf{Weather-free Prompt} & \textbf{Residual Prior} & PSNR & SSIM \\ 
        \midrule[0.5pt]
        \ding{51} & & & 31.07 & 0.9229  \\
        \ding{51} & \ding{51} & & 31.10 & 0.9244  \\
        \ding{51} &  & \ding{51} & 31.11 & 0.9243  \\
        \ding{51} & \ding{51} & \ding{51} & \textbf{31.18} & \textbf{0.9251}  \\
        \bottomrule[1pt]
    \end{tabular}
    }
    \label{tab:abl_EPM}
\end{table}

\textbf{Number of Learnable Vectors}. The input-conditional vector in C2P is of the size $N\times C$. To further study the impact of the learnable vectors for adverse weather removal, we initialize the vectors with different numbers (\emph{i.e.}, $N$), where ``0'' denotes the model without the vector. The results are summarized in Table~\ref{tab:prefixlen}. We can see that the performance boosts gradually as the number increases and achieves optimal performance across all degradation types when the vector length is set to 8. This is due to more vectors yielding better capacity for context understanding and weather-related prompt adaption. However, when we increase $N$ from 8 to 12, the model suffers from a performance decrease. This can be attributed to the overlarge vectors introducing unnecessary noisy information that overwhelms other important components. Therefore, we set $N=8$ in our experiments.

\textbf{Effect of EPM}. In this work, the EPM aims at leveraging constrained restoration priors to embed weather-free representations into the C2P, constituting a cyclic C2P to further refine the restoration process. To study its effect, we use the baseline model with initial C2P as the reference. As shown in Table~\ref{tab:abl_EPM}, the model replaces the visual prompt in the initial C2P with the weather-free prompt from the generated image in the 1st iteration can discover restoration-related information to improve the decoder features, thereby leading to higher scores. 

\begin{figure}[h]
    \centering  
    \includegraphics[width=\linewidth]{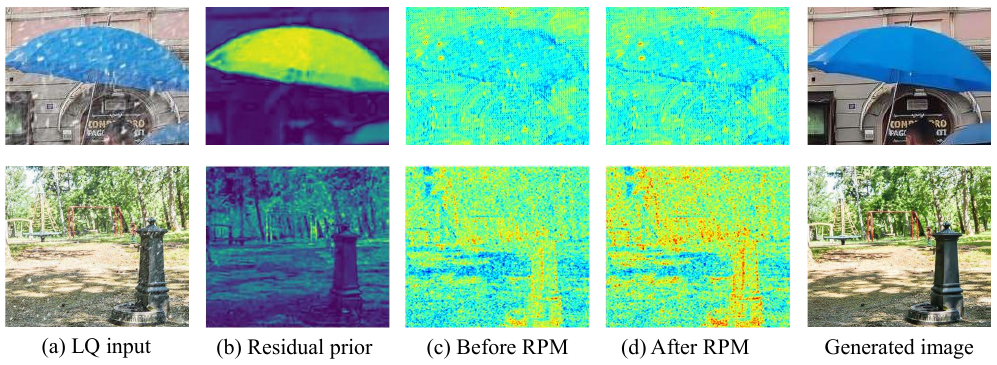}  
    \caption{Visualization of the feature maps before and after RPM.}
    \label{fig:abl_vis_rcp} 
\end{figure}

In CyclicPrompt, the residual prior is incorporated to enrich visual characteristics for better detail recovery along with EPM. We now present the experimental results to verify its impact. In Table~\ref{tab:abl_EPM}, we observe that it contributes to performance gains. When we combine it with the weather-free prompt, one can see that it assists the model well and enhances the performance notably. Moreover, in Fig.~\ref{fig:abl_vis_rcp}, we visualize the residual prior map and corresponding feature maps before and after the RPM. There are two observations: (1) the prior (Fig.~\ref{fig:abl_vis_rcp}(b)) is weather-invariant and allows expressing the contextualized information of scenes. (2) By modulating the decoder feature (Fig.~\ref{fig:abl_vis_rcp}(c)) with the prior, the model can effectively emphasize the textures and structures (Fig.~\ref{fig:abl_vis_rcp}(d)), improving the representation capacity for reconstruction. 

\textbf{Impact of Hypaerparameter $\lambda$}. We conduct six groups of experiments with the $\lambda$ value ranging from 0.1 to 4, respectively, where the results are illustrated in Table~\ref{tab:loss_iter2}. 
We can see that as $\lambda$ increases within the range of $\leq1$, the model performance demonstrates observable gains and converges when $\lambda=1$, which indicates that enhancing the impact of the second iteration contributes to improved model effectiveness. Besides, $\lambda > 1$ amplifies the impact of the second iteration beyond the first, where the model performance exhibits minimal fluctuations. Hence, we ultimately selected $\lambda=1$ in our experiments.

\begin{table}[t] 
\renewcommand\arraystretch{}
\caption{The Impact of the Hyperparameter $\lambda$}
\label{tab:loss_iter2}
\centering
\begin{tabular}{c c c c c c c }
\toprule
\textbf{$\lambda$}  & 0.1 & 0.5 & 1 & 2 & 3 & 4  \\ 
\midrule[0.5pt]
    PSNR & 31.06 & 31.10 & 31.18 & 31.17 & 31.18 & 31.18 \\
    SSIM & 0.9243 & 0.9245 & 0.9251 & 0.9253 & 0.9252 & 0.9243 \\
\bottomrule[1pt]
\end{tabular}
\end{table}

\textbf{Influence of Iteration Number}. The proposed CyclicPrompt contains two iterations that perform ``Prompt-Restore-Prompt''. To study the influence of the iteration number in UAWR, based on the second iteration, we further extract high-quality image features from the restoration results as restoration priors and perform refinement, thereby constructing model variants with 3 and 4 iterations, respectively. As quantitatively demonstrated in Table \ref{tab:iteration_num}, the PSNR metric exhibits insensitivity to the iteration number. While extending the model to 3 iterations yields a marginal SSIM improvement (0.9266 vs. 0.9251), this gain incurs a substantial computational overhead of about 90.5G FLOPs. Furthermore, the model with 4 iterations surprisingly degrades in SSIM with more FLOPs. Consequently, to maintain an optimal balance between efficiency and restoration quality, we set the iteration number to 2 as the final configuration.

\begin{table}[t] 
\renewcommand\arraystretch{}
\caption{The Influence of Iteration Number.}
\label{tab:iteration_num}
\centering
\begin{tabular}{c c c c c }
\toprule
Iteration Number & 	PSNR & SSIM	& Flops (G)\\ 
\midrule[0.5pt]
    2	& 31.18 & 0.9251 & 230.42 \\
    3	& 31.18 & 0.9266 & 320.90 \\
    4	& 31.18 & 0.9248 & 411.39 \\

\bottomrule[1pt]
\end{tabular}
\end{table}

\subsection{Discussion and Limitations}
\begin{figure}[h]
    \centering
    \includegraphics[width=\linewidth]{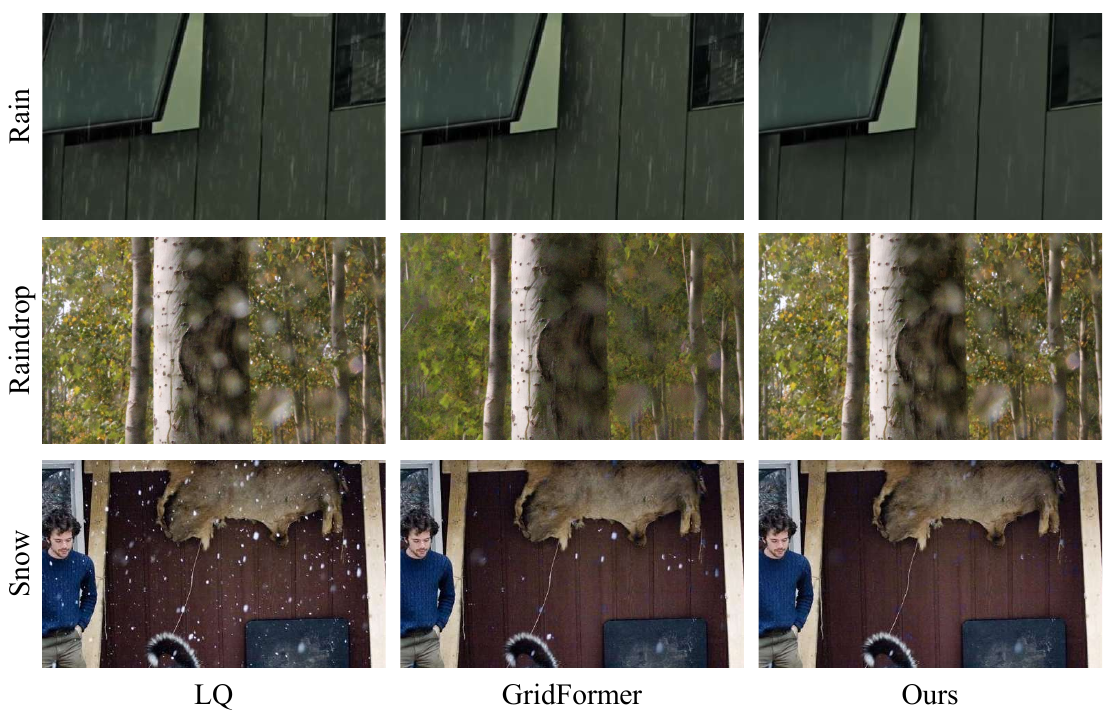} 
    \caption{Failure cases of our method under challenging scenarios.}
    \label{fig:failure_case}
\end{figure}

CyclicPrompt introduces a cyclic prompting framework that integrates C2P and EPM to constitute a cyclic prompting mechanism for UAWR. It involves a two-stage processing (``Prompt-Restore'' and ``Restore-Prompt''), which suffers from more computational burdens (Table~\ref{tab:realworld}). Despite its advantages, as shown in Fig.~\ref{fig:failure_case}, CyclicPrompt struggles to remove weather artifacts in regions where weather patterns and image textures become perceptually indistinguishable. For example, in the 1st row, the rain streaks are more like linear textures on the board, which can cause a misleading understanding of the scene context. Even so, we can see that our method still restores the images better than the recent SotA method GridFormer.

\section{Conclusion}
\label{sec:conclusion}
In this paper, we presented CyclicPrompt, a novel cyclic prompt learning framework for universal adverse weather removal. By integrating weather-specific knowledge, textual contexts, and constrained restoration priors, CyclicPrompt effectively addresses the challenges of handling diverse and complex weather degradations. Based on the proposed composite context prompt (C2P) and erase-and-paste mechanism (EPM), which together form a cyclic ``Prompt-Restore-Prompt'' pipeline. Extensive experimental results demonstrate the superiority of our CycliPrompt under various weather degradations.

\bibliographystyle{IEEEtran}
\bibliography{main}

\begin{thebibliography}{100}
\providecommand{\url}[1]{#1}
\csname url@samestyle\endcsname
\providecommand{\newblock}{\relax}
\providecommand{\bibinfo}[2]{#2}
\providecommand{\BIBentrySTDinterwordspacing}{\spaceskip=0pt\relax}
\providecommand{\BIBentryALTinterwordstretchfactor}{4}
\providecommand{\BIBentryALTinterwordspacing}{\spaceskip=\fontdimen2\font plus
\BIBentryALTinterwordstretchfactor\fontdimen3\font minus \fontdimen4\font\relax}
\providecommand{\BIBforeignlanguage}[2]{{%
\expandafter\ifx\csname l@#1\endcsname\relax
\typeout{** WARNING: IEEEtran.bst: No hyphenation pattern has been}%
\typeout{** loaded for the language `#1'. Using the pattern for}%
\typeout{** the default language instead.}%
\else
\language=\csname l@#1\endcsname
\fi
#2}}
\providecommand{\BIBdecl}{\relax}
\BIBdecl

\bibitem{Ithaca365}
C.~A. Diaz-Ruiz, Y.~Xia, Y.~You, J.~Nino, J.~Chen, J.~Monica, X.~Chen, K.~Luo, Y.~Wang, M.~Emond \emph{et~al.}, ``Ithaca365: Dataset and driving perception under repeated and challenging weather conditions,'' in \emph{Proceedings of the IEEE/CVF conference on computer vision and pattern recognition}, 2022, pp. 21\,383--21\,392.

\bibitem{reddy2022master}
N.~Reddy, A.~Singhal, A.~Kumar, M.~Baktashmotlagh, and C.~Arora, ``Master of all: Simultaneous generalization of urban-scene segmentation to all adverse weather conditions,'' in \emph{Proceedings of the European conference on computer vision}, 2022, pp. 51--69.

\bibitem{weatherstream}
H.~Zhang, Y.~Ba, E.~Yang, V.~Mehra, B.~Gella, A.~Suzuki, A.~Pfahnl, C.~C. Chandrappa, A.~Wong, and A.~Kadambi, ``Weatherstream: Light transport automation of single image deweathering,'' in \emph{Proceedings of the IEEE/CVF conference on computer vision and pattern recognition}, 2023, pp. 13\,499--13\,509.

\bibitem{li2023dilated}
Y.~Li, J.~Lu, H.~Chen, X.~Wu, and X.~Chen, ``Dilated convolutional transformer for high-quality image deraining,'' in \emph{Proceedings of the IEEE/CVF conference on computer vision and pattern recognition}, 2023, pp. 4198--4206.

\bibitem{chen2023learning}
X.~Chen, H.~Li, M.~Li, and J.~Pan, ``Learning a sparse transformer network for effective image deraining,'' in \emph{Proceedings of the IEEE/CVF conference on computer vision and pattern recognition}, 2023, pp. 5896--5905.

\bibitem{wei2021deraincyclegan}
Y.~Wei, Z.~Zhang, Y.~Wang, M.~Xu, Y.~Yang, S.~Yan, and M.~Wang, ``Deraincyclegan: Rain attentive cyclegan for single image deraining and rainmaking,'' \emph{IEEE transactions on image processing}, vol.~30, pp. 4788--4801, 2021.

\bibitem{dehaze-sam}
Z.~Jin, S.~Chen, Y.~Chen, Z.~Xu, and H.~Feng, ``Let segment anything help image dehaze,'' \emph{arXiv preprint arXiv:2306.15870}, 2023.

\bibitem{guo2022image}
C.-L. Guo, Q.~Yan, S.~Anwar, R.~Cong, W.~Ren, and C.~Li, ``Image dehazing transformer with transmission-aware 3d position embedding,'' in \emph{Proceedings of the IEEE/CVF conference on computer vision and pattern recognition}, 2022, pp. 5812--5820.

\bibitem{Dataset-Snow100K}
Y.-F. Liu, D.-W. Jaw, S.-C. Huang, and J.-N. Hwang, ``Desnownet: Context-aware deep network for snow removal,'' \emph{IEEE transactions on image processing}, vol.~27, no.~6, pp. 3064--3073, 2018.

\bibitem{Restormer}
S.~W. Zamir, A.~Arora, S.~Khan, M.~Hayat, F.~S. Khan, and M.-H. Yang, ``Restormer: Efficient transformer for high-resolution image restoration,'' in \emph{Proceedings of the IEEE/CVF conference on computer vision and pattern recognition}, 2022, pp. 5728--5739.

\bibitem{Uformer}
Z.~Wang, X.~Cun, J.~Bao, W.~Zhou, J.~Liu, and H.~Li, ``Uformer: A general u-shaped transformer for image restoration,'' in \emph{Proceedings of the IEEE/CVF conference on computer vision and pattern recognition}, 2022, pp. 17\,683--17\,693.

\bibitem{GRL}
Y.~Li, Y.~Fan, X.~Xiang, D.~Demandolx, R.~Ranjan, R.~Timofte, and L.~Van~Gool, ``Efficient and explicit modelling of image hierarchies for image restoration,'' in \emph{Proceedings of the IEEE/CVF conference on computer vision and pattern recognition}, 2023, pp. 18\,278--18\,289.

\bibitem{chen2021ipt}
H.~Chen, Y.~Wang, T.~Guo, C.~Xu, Y.~Deng, Z.~Liu, S.~Ma, C.~Xu, C.~Xu, and W.~Gao, ``Pre-trained image processing transformer,'' in \emph{Proceedings of the IEEE/CVF conference on computer vision and pattern recognition}, 2021, pp. 12\,299--12\,310.

\bibitem{Dataset-AllWeather}
R.~Li, R.~T. Tan, and L.-F. Cheong, ``All in one bad weather removal using architectural search,'' in \emph{Proceedings of the IEEE/CVF conference on computer vision and pattern recognition}, 2020, pp. 3175--3185.

\bibitem{TransWeather}
J.~M.~J. Valanarasu, R.~Yasarla, and V.~M. Patel, ``Transweather: Transformer-based restoration of images degraded by adverse weather conditions,'' in \emph{Proceedings of the IEEE/CVF conference on computer vision and pattern recognition}, 2022, pp. 2353--2363.

\bibitem{ADFSD}
D.~Park, B.~H. Lee, and S.~Y. Chun, ``All-in-one image restoration for unknown degradations using adaptive discriminative filters for specific degradations,'' in \emph{Proceedings of the IEEE/CVF conference on computer vision and pattern recognition}, 2023, pp. 5815--5824.

\bibitem{TUM_chen}
W.-T. Chen, Z.-K. Huang, C.-C. Tsai, H.-H. Yang, J.-J. Ding, and S.-Y. Kuo, ``Learning multiple adverse weather removal via two-stage knowledge learning and multi-contrastive regularization: Toward a unified model,'' in \emph{Proceedings of the IEEE/CVF conference on computer vision and pattern recognition}, 2022, pp. 17\,653--17\,662.

\bibitem{domainMPR}
P.~W. Patil, S.~Gupta, S.~Rana, S.~Venkatesh, and S.~Murala, ``Multi-weather image restoration via domain translation,'' in \emph{Proceedings of the IEEE/CVF international conference on computer vision}, 2023, pp. 21\,696--21\,705.

\bibitem{AWRCP}
T.~Ye, S.~Chen, J.~Bai, J.~Shi, C.~Xue, J.~Jiang, J.~Yin, E.~Chen, and Y.~Liu, ``Adverse weather removal with codebook priors,'' in \emph{Proceedings of the IEEE/CVF international conference on computer vision}, 2023, pp. 12\,653--12\,664.

\bibitem{CLIP}
A.~Radford, J.~W. Kim, C.~Hallacy, A.~Ramesh, G.~Goh, S.~Agarwal, G.~Sastry, A.~Askell, P.~Mishkin, J.~Clark \emph{et~al.}, ``Learning transferable visual models from natural language supervision,'' in \emph{Proceedings of the International conference on machine learning}, 2021, pp. 8748--8763.

\bibitem{LISA}
X.~Lai, Z.~Tian, Y.~Chen, Y.~Li, Y.~Yuan, S.~Liu, and J.~Jia, ``Lisa: Reasoning segmentation via large language model,'' in \emph{Proceedings of the IEEE/CVF conference on computer vision and pattern recognition}, 2024, pp. 9579--9589.

\bibitem{SD}
R.~Rombach, A.~Blattmann, D.~Lorenz, P.~Esser, and B.~Ommer, ``High-resolution image synthesis with latent diffusion models,'' in \emph{Proceedings of the IEEE/CVF conference on computer vision and pattern recognition}, 2022, pp. 10\,684--10\,695.

\bibitem{DA-CLIP}
Z.~Luo, F.~K. Gustafsson, Z.~Zhao, J.~Sj{\"o}lund, and T.~B. Sch{\"o}n, ``Controlling vision-language models for universal image restoration,'' in \emph{Proceedings of the international conference on learning representations}, 2024.

\bibitem{CVPR24-Languagedriven}
H.~Yang, L.~Pan, Y.~Yang, and W.~Liang, ``Language-driven all-in-one adverse weather removal,'' in \emph{Proceedings of the IEEE/CVF conference on computer vision and pattern recognition}, 2024, pp. 24\,902--24\,912.

\bibitem{tan_tip24}
Z.~Tan, Y.~Wu, Q.~Liu, Q.~Chu, L.~Lu, J.~Ye, and N.~Yu, ``Exploring the application of large-scale pre-trained models on adverse weather removal,'' \emph{IEEE transactions on image processing}, 2024.

\bibitem{T3-DiffWeather}
S.~Chen, T.~Ye, K.~Zhang, Z.~Xing, Y.~Lin, and L.~Zhu, ``Teaching tailored to talent: Adverse weather restoration via prompt pool and depth-anything constraint,'' in \emph{Proceedings of the European conference on computer vision}, 2024, pp. 95--115.

\bibitem{WeatherDiff}
O.~{\"O}zdenizci and R.~Legenstein, ``Restoring vision in adverse weather conditions with patch-based denoising diffusion models,'' \emph{IEEE transactions on pattern analysis and machine intelligence}, 2023.

\bibitem{Dataset-RainDS}
R.~Quan, X.~Yu, Y.~Liang, and Y.~Yang, ``Removing raindrops and rain streaks in one go,'' in \emph{Proceedings of the IEEE/CVF conference on computer vision and pattern recognition}, 2021, pp. 9147--9156.

\bibitem{Dataset-Raindrop}
R.~Qian, R.~T. Tan, W.~Yang, J.~Su, and J.~Liu, ``Attentive generative adversarial network for raindrop removal from a single image,'' in \emph{Proceedings of the IEEE conference on computer vision and pattern recognition}, 2018, pp. 2482--2491.

\bibitem{chen2021all}
W.-T. Chen, H.-Y. Fang, C.-L. Hsieh, C.-C. Tsai, I.~Chen, J.-J. Ding, S.-Y. Kuo \emph{et~al.}, ``All snow removed: Single image desnowing algorithm using hierarchical dual-tree complex wavelet representation and contradict channel loss,'' in \emph{Proceedings of the IEEE/CVF international conference on computer vision}, 2021, pp. 4196--4205.

\bibitem{DDMSNet}
K.~Zhang, R.~Li, Y.~Yu, W.~Luo, and C.~Li, ``Deep dense multi-scale network for snow removal using semantic and depth priors,'' \emph{IEEE transactions on image processing}, vol.~30, pp. 7419--7431, 2021.

\bibitem{wu2023ridcp}
R.-Q. Wu, Z.-P. Duan, C.-L. Guo, Z.~Chai, and C.~Li, ``Ridcp: Revitalizing real image dehazing via high-quality codebook priors,'' in \emph{Proceedings of the IEEE/CVF conference on computer vision and pattern recognition}, 2023, pp. 22\,282--22\,291.

\bibitem{HRGAN}
R.~Li, L.-F. Cheong, and R.~T. Tan, ``Heavy rain image restoration: Integrating physics model and conditional adversarial learning,'' in \emph{Proceedings of the IEEE/CVF conference on computer vision and pattern recognition}, 2019, pp. 1633--1642.

\bibitem{Pix2Pix}
P.~Isola, J.-Y. Zhu, T.~Zhou, and A.~A. Efros, ``Image-to-image translation with conditional adversarial networks,'' in \emph{Proceedings of the IEEE conference on computer vision and pattern recognition}, 2017, pp. 1125--1134.

\bibitem{fu2017clearing}
X.~Fu, J.~Huang, X.~Ding, Y.~Liao, and J.~Paisley, ``Clearing the skies: A deep network architecture for single-image rain removal,'' \emph{IEEE transactions on image processing}, vol.~26, no.~6, pp. 2944--2956, 2017.

\bibitem{Dataset-Rain100H}
W.~Yang, R.~T. Tan, J.~Feng, J.~Liu, Z.~Guo, and S.~Yan, ``Deep joint rain detection and removal from a single image,'' in \emph{Proceedings of the IEEE conference on computer vision and pattern recognition}, 2017, pp. 1357--1366.

\bibitem{li2018recurrent}
X.~Li, J.~Wu, Z.~Lin, H.~Liu, and H.~Zha, ``Recurrent squeeze-and-excitation context aggregation net for single image deraining,'' in \emph{Proceedings of the European conference on computer vision (ECCV)}, 2018, pp. 254--269.

\bibitem{RainGAN-ID-CGAN}
H.~Zhang, V.~Sindagi, and V.~M. Patel, ``Image de-raining using a conditional generative adversarial network,'' \emph{IEEE transactions on circuits and systems for video technology}, vol.~30, no.~11, pp. 3943--3956, 2019.

\bibitem{deng2020detail}
S.~Deng, M.~Wei, J.~Wang, Y.~Feng, L.~Liang, H.~Xie, F.~L. Wang, and M.~Wang, ``Detail-recovery image deraining via context aggregation networks,'' in \emph{Proceedings of the IEEE/CVF conference on computer vision and pattern recognition}, 2020, pp. 14\,560--14\,569.

\bibitem{HazeCNN-GCANet}
D.~Chen, M.~He, Q.~Fan, J.~Liao, L.~Zhang, D.~Hou, L.~Yuan, and G.~Hua, ``Gated context aggregation network for image dehazing and deraining,'' in \emph{Proceedings of the IEEE/CVF winter conference on applications of computer vision}.\hskip 1em plus 0.5em minus 0.4em\relax IEEE, 2019, pp. 1375--1383.

\bibitem{RainCNN-PReNet}
D.~Ren, W.~Zuo, Q.~Hu, P.~Zhu, and D.~Meng, ``Progressive image deraining networks: A better and simpler baseline,'' in \emph{Proceedings of the IEEE/CVF conference on computer vision and pattern recognition}, 2019, pp. 3937--3946.

\bibitem{Rain-Rcp}
Q.~Yi, J.~Li, Q.~Dai, F.~Fang, G.~Zhang, and T.~Zeng, ``Structure-preserving deraining with residue channel prior guidance,'' in \emph{Proceedings of the IEEE/CVF international conference on computer vision}, 2021, pp. 4238--4247.

\bibitem{wang2020model}
H.~Wang, Q.~Xie, Q.~Zhao, and D.~Meng, ``A model-driven deep neural network for single image rain removal,'' in \emph{Proceedings of the IEEE/CVF conference on computer vision and pattern recognition}, 2020, pp. 3103--3112.

\bibitem{pan2020physics}
J.~Pan, J.~Dong, Y.~Liu, J.~Zhang, J.~Ren, J.~Tang, Y.-W. Tai, and M.-H. Yang, ``Physics-based generative adversarial models for image restoration and beyond,'' \emph{IEEE transactions on pattern analysis and machine intelligence}, vol.~43, no.~7, pp. 2449--2462, 2020.

\bibitem{yu2023both}
C.~Yu, S.~Chen, Y.~Chang, Y.~Song, and L.~Yan, ``Both diverse and realism matter: Physical attribute and style alignment for rainy image generation,'' in \emph{Proceedings of the IEEE/CVF international conference on computer vision}, 2023, pp. 12\,387--12\,397.

\bibitem{li2018robust}
R.~Li, R.~T. Tan, and L.-F. Cheong, ``Robust optical flow in rainy scenes,'' in \emph{Proceedings of the European conference on computer vision}, 2018, pp. 288--304.

\bibitem{RainTransformer-IDT}
J.~Xiao, X.~Fu, A.~Liu, F.~Wu, and Z.-J. Zha, ``Image de-raining transformer,'' \emph{IEEE transactions on pattern analysis and machine intelligence}, 2022.

\bibitem{wang2023multi}
Q.~Wang, K.~Jiang, Z.~Wang, W.~Ren, J.~Zhang, and C.-W. Lin, ``Multi-scale fusion and decomposition network for single image deraining,'' \emph{IEEE transactions on image processing}, vol.~33, pp. 191--204, 2023.

\bibitem{quan2019deep}
Y.~Quan, S.~Deng, Y.~Chen, and H.~Ji, ``Deep learning for seeing through window with raindrops,'' in \emph{Proceedings of the IEEE/CVF international conference on computer vision}, 2019, pp. 2463--2471.

\bibitem{shao2021uncertainty}
M.-W. Shao, L.~Li, D.-Y. Meng, and W.-M. Zuo, ``Uncertainty guided multi-scale attention network for raindrop removal from a single image,'' \emph{IEEE transactions on image processing}, vol.~30, pp. 4828--4839, 2021.

\bibitem{zhang2021dual}
K.~Zhang, D.~Li, W.~Luo, and W.~Ren, ``Dual attention-in-attention model for joint rain streak and raindrop removal,'' \emph{IEEE transactions on image processing}, vol.~30, pp. 7608--7619, 2021.

\bibitem{cai2016dehazenet}
B.~Cai, X.~Xu, K.~Jia, C.~Qing, and D.~Tao, ``Dehazenet: An end-to-end system for single image haze removal,'' \emph{IEEE transactions on image processing}, vol.~25, no.~11, pp. 5187--5198, 2016.

\bibitem{li2017aod}
B.~Li, X.~Peng, Z.~Wang, J.~Xu, and D.~Feng, ``Aod-net: All-in-one dehazing network,'' in \emph{Proceedings of the IEEE international conference on computer vision}, 2017, pp. 4770--4778.

\bibitem{zhang2018densely}
H.~Zhang and V.~M. Patel, ``Densely connected pyramid dehazing network,'' in \emph{Proceedings of the IEEE conference on computer vision and pattern recognition}, 2018, pp. 3194--3203.

\bibitem{HazeGAN-PMHLD}
W.-T. Chen, H.-Y. Fang, J.-J. Ding, and S.-Y. Kuo, ``Pmhld: Patch map-based hybrid learning dehazenet for single image haze removal,'' \emph{IEEE transactions on image processing}, vol.~29, pp. 6773--6788, 2020.

\bibitem{qin2020ffa}
X.~Qin, Z.~Wang, Y.~Bai, X.~Xie, and H.~Jia, ``Ffa-net: Feature fusion attention network for single image dehazing,'' in \emph{Proceedings of the AAAI conference on artificial intelligence}, vol.~34, no.~07, 2020, pp. 11\,908--11\,915.

\bibitem{dong2020multi}
H.~Dong, J.~Pan, L.~Xiang, Z.~Hu, X.~Zhang, F.~Wang, and M.-H. Yang, ``Multi-scale boosted dehazing network with dense feature fusion,'' in \emph{Proceedings of the IEEE/CVF conference on computer vision and pattern recognition}, 2020, pp. 2157--2167.

\bibitem{yang2022self}
Y.~Yang, C.~Wang, R.~Liu, L.~Zhang, X.~Guo, and D.~Tao, ``Self-augmented unpaired image dehazing via density and depth decomposition,'' in \emph{Proceedings of the IEEE/CVF conference on computer vision and pattern recognition}, 2022, pp. 2037--2046.

\bibitem{wang2024selfpromer}
C.~Wang, J.~Pan, W.~Lin, J.~Dong, W.~Wang, and X.-M. Wu, ``Selfpromer: Self-prompt dehazing transformers with depth-consistency,'' in \emph{Proceedings of the AAAI conference on artificial intelligence}, 2024, pp. 5327--5335.

\bibitem{zhang2024depth}
Y.~Zhang, S.~Zhou, and H.~Li, ``Depth information assisted collaborative mutual promotion network for single image dehazing,'' in \emph{Proceedings of the IEEE/CVF conference on computer vision and pattern recognition}, 2024, pp. 2846--2855.

\bibitem{ye2024learning}
T.~Ye, S.~Chen, W.~Chai, Z.~Xing, J.~Qin, G.~Lin, and L.~Zhu, ``Learning diffusion texture priors for image restoration,'' in \emph{Proceedings of the IEEE/CVF conference on computer vision and pattern recognition}, 2024, pp. 2524--2534.

\bibitem{JSTASR}
W.-T. Chen, H.-Y. Fang, J.-J. Ding, C.-C. Tsai, and S.-Y. Kuo, ``Jstasr: Joint size and transparency-aware snow removal algorithm based on modified partial convolution and veiling effect removal,'' in \emph{Proceedings of the European conference on computer vision}, 2020, pp. 754--770.

\bibitem{liang2022drt}
Y.~Liang, S.~Anwar, and Y.~Liu, ``Drt: A lightweight single image deraining recursive transformer,'' in \emph{Proceedings of the IEEE/CVF conference on computer vision and pattern recognition}, 2022, pp. 589--598.

\bibitem{chen2023uncertainty}
S.~Chen, T.~Ye, C.~Xue, H.~Chen, Y.~Liu, E.~Chen, and L.~Zhu, ``Uncertainty-driven dynamic degradation perceiving and background modeling for efficient single image desnowing,'' in \emph{Proceedings of the ACM international conference on multimedia}, 2023, pp. 4269--4280.

\bibitem{chen2023cplformer}
S.~Chen, T.~Ye, Y.~Liu, J.~Bai, H.~Chen, Y.~Lin, J.~Shi, and E.~Chen, ``Cplformer: Cross-scale prototype learning transformer for image snow removal,'' in \emph{Proceedings of the ACM international conference on multimedia}, 2023, pp. 4228--4239.

\bibitem{zhang2023hcsd}
T.~Zhang, N.~Jiang, H.~Wu, K.~Zhang, Y.~Niu, and T.~Zhao, ``Hcsd-net: Single image desnowing with color space transformation,'' in \emph{Proceedings of the ACM international conference on multimedia}, 2023, pp. 8125--8133.

\bibitem{AirNet}
B.~Li, X.~Liu, P.~Hu, Z.~Wu, J.~Lv, and X.~Peng, ``All-in-one image restoration for unknown corruption,'' in \emph{Proceedings of the IEEE/CVF conference on computer vision and pattern recognition}, 2022, pp. 17\,452--17\,462.

\bibitem{WGWS}
Y.~Zhu, T.~Wang, X.~Fu, X.~Yang, X.~Guo, J.~Dai, Y.~Qiao, and X.~Hu, ``Learning weather-general and weather-specific features for image restoration under multiple adverse weather conditions,'' in \emph{Proceedings of the IEEE/CVF conference on computer vision and pattern recognition}, 2023, pp. 21\,747--21\,758.

\bibitem{VQGAN}
P.~Esser, R.~Rombach, and B.~Ommer, ``Taming transformers for high-resolution image synthesis,'' in \emph{Proceedings of the IEEE/CVF conference on computer vision and pattern recognition}, 2021, pp. 12\,873--12\,883.

\bibitem{DDPM}
J.~Ho, A.~Jain, and P.~Abbeel, ``Denoising diffusion probabilistic models,'' \emph{Advances in neural information processing systems}, pp. 6840--6851, 2020.

\bibitem{bert}
J.~Devlin, ``Bert: Pre-training of deep bidirectional transformers for language understanding,'' \emph{arXiv preprint arXiv:1810.04805}, 2018.

\bibitem{GPT}
A.~Radford, J.~Wu, R.~Child, D.~Luan, D.~Amodei, I.~Sutskever \emph{et~al.}, ``Language models are unsupervised multitask learners,'' \emph{OpenAI blog}, vol.~1, no.~8, p.~9, 2019.

\bibitem{lester2021power}
B.~Lester, R.~Al-Rfou, and N.~Constant, ``The power of scale for parameter-efficient prompt tuning,'' \emph{arXiv preprint arXiv:2104.08691}, 2021.

\bibitem{liu2021p}
X.~Liu, K.~Ji, Y.~Fu, W.~L. Tam, Z.~Du, Z.~Yang, and J.~Tang, ``P-tuning v2: Prompt tuning can be comparable to fine-tuning universally across scales and tasks,'' \emph{arXiv preprint arXiv:2110.07602}, 2021.

\bibitem{jia2021scaling}
C.~Jia, Y.~Yang, Y.~Xia, Y.-T. Chen, Z.~Parekh, H.~Pham, Q.~Le, Y.-H. Sung, Z.~Li, and T.~Duerig, ``Scaling up visual and vision-language representation learning with noisy text supervision,'' in \emph{Proceedings of the international conference on machine learning}.\hskip 1em plus 0.5em minus 0.4em\relax PMLR, 2021, pp. 4904--4916.

\bibitem{jia2022visual}
M.~Jia, L.~Tang, B.-C. Chen, C.~Cardie, S.~Belongie, B.~Hariharan, and S.-N. Lim, ``Visual prompt tuning,'' in \emph{Proceedings of the European conference on computer vision}, 2022, pp. 709--727.

\bibitem{bahng2022visual}
H.~Bahng, A.~Jahanian, S.~Sankaranarayanan, and P.~Isola, ``Visual prompting: Modifying pixel space to adapt pre-trained models,'' \emph{arXiv preprint arXiv:2203.17274}, 2022.

\bibitem{Coop}
K.~Zhou, J.~Yang, C.~C. Loy, and Z.~Liu, ``Learning to prompt for vision-language models,'' \emph{International journal of computer vision}, vol. 130, no.~9, pp. 2337--2348, 2022.

\bibitem{cocoop}
K.~\vspace{0mm}Zhou, J.~Yang, C.~C. Loy, and Z.~Liu, ``Conditional prompt learning for vision-language models,'' in \emph{Proceedings of the IEEE/CVF conference on computer vision and pattern recognition}, 2022, pp. 16\,816--16\,825.

\bibitem{CLIP-Lit}
Z.~Liang, C.~Li, S.~Zhou, R.~Feng, and C.~C. Loy, ``Iterative prompt learning for unsupervised backlit image enhancement,'' in \emph{Proceedings of the IEEE/CVF international conference on computer vision}, 2023, pp. 8094--8103.

\bibitem{Seesr_CVPR24}
R.~Wu, T.~Yang, L.~Sun, Z.~Zhang, S.~Li, and L.~Zhang, ``Seesr: Towards semantics-aware real-world image super-resolution,'' in \emph{Proceedings of the IEEE/CVF conference on computer vision and pattern recognition}, 2024, pp. 25\,456--25\,467.

\bibitem{BLIP}
J.~Li, D.~Li, C.~Xiong, and S.~Hoi, ``Blip: Bootstrapping language-image pre-training for unified vision-language understanding and generation,'' in \emph{Proceedings of the international conference on machine learning}, 2022, pp. 12\,888--12\,900.

\bibitem{PromptIR}
V.~Potlapalli, S.~W. Zamir, S.~Khan, and F.~Khan, ``Promptir: Prompting for all-in-one image restoration,'' in \emph{Proceedings of the advances in neural information processing systems}, 2023.

\bibitem{CVHSSR}
W.~Zou, H.~Gao, L.~Chen, Y.~Zhang, M.~Jiang, Z.~Yu, and M.~Tan, ``Cross-view hierarchy network for stereo image super-resolution,'' in \emph{Proceedings of the IEEE/CVF conference on computer vision and pattern recognition}, 2023, pp. 1396--1405.

\bibitem{PCNet}
K.~Jiang, Z.~Wang, P.~Yi, C.~Chen, Z.~Wang, X.~Wang, J.~Jiang, and C.-W. Lin, ``Rain-free and residue hand-in-hand: A progressive coupled network for real-time image deraining,'' \emph{IEEE transactions on image processing}, vol.~30, pp. 7404--7418, 2021.

\bibitem{MPRNet}
S.~W. Zamir, A.~Arora, S.~Khan, M.~Hayat, F.~S. Khan, M.-H. Yang, and L.~Shao, ``Multi-stage progressive image restoration,'' in \emph{Proceedings of the IEEE/CVF conference on computer vision and pattern recognition}, 2021, pp. 14\,821--14\,831.

\bibitem{Art}
G.~Wu, J.~Jiang, K.~Jiang, and X.~Liu, ``Harmony in diversity: Improving all-in-one image restoration via multi-task collaboration,'' in \emph{Proceedings of the 32nd ACM international conference on multimedia}, 2024, pp. 6015--6023.

\bibitem{wang2024gridformer}
T.~Wang, K.~Zhang, Z.~Shao, W.~Luo, B.~Stenger, T.~Lu, T.-K. Kim, W.~Liu, and H.~Li, ``Gridformer: Residual dense transformer with grid structure for image restoration in adverse weather conditions,'' \emph{International journal of computer vision}, vol. 132, pp. 4541--4563, 2024.

\bibitem{Dataset-SPA}
T.~Wang, X.~Yang, K.~Xu, S.~Chen, Q.~Zhang, and R.~W. Lau, ``Spatial attentive single-image deraining with a high quality real rain dataset,'' in \emph{Proceedings of the IEEE/CVF conference on computer vision and pattern recognition}, 2019, pp. 12\,270--12\,279.

\bibitem{Dataset-Rain100L}
F.~Yang, H.~Yang, J.~Fu, H.~Lu, and B.~Guo, ``Learning texture transformer network for image super-resolution,'' in \emph{Proceedings of the IEEE/CVF conference on computer vision and pattern recognition}, 2020, pp. 5791--5800.

\bibitem{Dataset-RESIDE-OTS}
B.~Li, W.~Ren, D.~Fu, D.~Tao, D.~Feng, W.~Zeng, and Z.~Wang, ``Benchmarking single-image dehazing and beyond,'' \emph{IEEE transactions on image processing}, vol.~28, no.~1, pp. 492--505, 2018.

\bibitem{Dataset-BSD400}
P.~Arbelaez, M.~Maire, C.~Fowlkes, and J.~Malik, ``Contour detection and hierarchical image segmentation,'' \emph{IEEE transactions on pattern analysis and machine intelligence}, vol.~33, no.~5, pp. 898--916, 2010.

\bibitem{Dataset-GOPRO}
S.~Nah, T.~Hyun~Kim, and K.~Mu~Lee, ``Deep multi-scale convolutional neural network for dynamic scene deblurring,'' in \emph{Proceedings of the IEEE conference on computer vision and pattern recognition}, 2017, pp. 3883--3891.

\bibitem{Dataset-LOL}
C.~Wei, W.~Wang, W.~Yang, and J.~Liu, ``Deep retinex decomposition for low-light enhancement,'' in \emph{Proceedings of the British Machine Vision Conference}, 2018.

\bibitem{HINet}
L.~Chen, X.~Lu, J.~Zhang, X.~Chu, and C.~Chen, ``Hinet: Half instance normalization network for image restoration,'' in \emph{Proceedings of the IEEE/CVF conference on computer vision and pattern recognition}, 2021, pp. 182--192.

\bibitem{DGUNet}
C.~Mou, Q.~Wang, and J.~Zhang, ``Deep generalized unfolding networks for image restoration,'' in \emph{Proceedings of the IEEE/CVF conference on computer vision and pattern recognition}, 2022, pp. 17\,399--17\,410.

\bibitem{MIRNetV2}
W.~Yu, M.~Luo, P.~Zhou, C.~Si, Y.~Zhou, X.~Wang, J.~Feng, and S.~Yan, ``Metaformer is actually what you need for vision,'' in \emph{Proceedings of the IEEE/CVF conference on computer vision and pattern recognition}, 2022, pp. 10\,819--10\,829.

\bibitem{SwinIR}
J.~Liang, J.~Cao, G.~Sun, K.~Zhang, L.~Van~Gool, and R.~Timofte, ``Swinir: Image restoration using swin transformer,'' in \emph{Proceedings of the IEEE/CVF international conference on computer vision}, 2021, pp. 1833--1844.

\bibitem{NAFNet}
L.~Chen, X.~Chu, X.~Zhang, and J.~Sun, ``Simple baselines for image restoration,'' in \emph{Proceedings of the European conference on computer vision}, 2022.

\bibitem{DL}
Q.~Fan, D.~Chen, L.~Yuan, G.~Hua, N.~Yu, and B.~Chen, ``A general decoupled learning framework for parameterized image operators,'' \emph{IEEE transactions on pattern analysis and machine intelligence}, vol.~43, no.~1, pp. 33--47, 2019.

\bibitem{TAPE}
L.~Liu, L.~Xie, X.~Zhang, S.~Yuan, X.~Chen, W.~Zhou, H.~Li, and Q.~Tian, ``Tape: Task-agnostic prior embedding for image restoration,'' in \emph{Proceedings of the European Conference on Computer Vision}, 2022, pp. 447--464.

\bibitem{IDR}
J.~Zhang, J.~Huang, M.~Yao, Z.~Yang, H.~Yu, M.~Zhou, and F.~Zhao, ``Ingredient-oriented multi-degradation learning for image restoration,'' in \emph{Proceedings of the IEEE/CVF conference on computer vision and pattern recognition}, 2023, pp. 5825--5835.

\bibitem{InstructIR}
M.~V. Conde, G.~Geigle, and R.~Timofte, ``Instructir: High-quality image restoration following human instructions,'' in \emph{Proceedings of the European conference on computer vision}, 2024.

\bibitem{NDR}
M.~Yao, R.~Xu, Y.~Guan, J.~Huang, and Z.~Xiong, ``Neural degradation representation learning for all-in-one image restoration,'' \emph{IEEE transactions on image processing}, 2024.

\bibitem{zheng2024selective}
D.~Zheng, X.-M. Wu, S.~Yang, J.~Zhang, J.-F. Hu, and W.-S. Zheng, ``Selective hourglass mapping for universal image restoration based on diffusion model,'' in \emph{Proceedings of the IEEE/CVF conference on computer vision and pattern recognition}, 2024, pp. 25\,445--25\,455.

\bibitem{ke2021musiq}
J.~Ke, Q.~Wang, Y.~Wang, P.~Milanfar, and F.~Yang, ``Musiq: Multi-scale image quality transformer,'' in \emph{Proceedings of the IEEE/CVF international conference on computer vision}, 2021, pp. 5148--5157.

\bibitem{yang2022maniqa}
S.~Yang, T.~Wu, S.~Shi, S.~Lao, Y.~Gong, M.~Cao, J.~Wang, and Y.~Yang, ``Maniqa: Multi-dimension attention network for no-reference image quality assessment,'' in \emph{Proceedings of the IEEE/CVF conference on computer vision and pattern recognition}, 2022, pp. 1191--1200.

\bibitem{mittal2012niqe}
A.~Mittal, R.~Soundararajan, and A.~C. Bovik, ``Making a “completely blind” image quality analyzer,'' \emph{IEEE signal processing letters}, vol.~20, no.~3, pp. 209--212, 2012.

\end{thebibliography}

\end{document}